Review article

# Dataset creation for supervised deep learning-based analysis of microscopic images – review of important considerations and recommendations

Christof A. Bertram [1], Viktoria Weiss [1], Jonas Ammeling [2], F. Maria Schabel [1], Taryn A. Donovan [3], Frauke Wilm [4,5], Christian Marzahl [6], Katharina Breininger [7], Marc Aubreville [8]

[1] University of Veterinary Medicine Vienna, Vienna, Austria

[2] Technische Hochschule Ingolstadt, Ingolstadt, Germany

[3] The Schwarzman Animal Medical Center, New York, NY, United States of America

[4] Friedrich-Alexander-Universität Erlangen-Nürnberg, Erlangen, Germany

[5] Mira Vision Microscopy GmbH, Göppingen, Germany

[6] Gestalt Diagnostics, Spokane, WA, United States of America

[7] Julius-Maximilians-Universität Würzburg

[8] Flensburg University of Applied Sciences, Flensburg, Germany

**Corresponding author:**

Christof Bertram, Institute of Pathology, University of Veterinary Medicine Vienna, Veterinärplatz 1, 1210 Vienna, Austria. Email: Christof.bertram@vetmeduni.ac.at

**Abstract:**

Supervised deep learning (DL) receives great interest for automated analysis of microscopic images with an increasing body of literature supporting its potential. The development and validation of those DL models relies heavily on the availability of high-quality, large-scale datasets. However, creating such datasets is a complex and resource-intensive process, often hindered by challenges such as time constraints, domain variability, and risks of bias in image collection and label creation. This review provides a comprehensive guide to the critical steps in dataset creation, including: 1) image acquisition, 2) selection of annotation software and 3) annotation creation. In addition to ensuring a sufficiently large number of images, it is crucial to address sources of image variability (domain shifts) – such as those related to slide preparation and digitization – that could lead to algorithmic errors if not adequately represented in the training data. Key quality criteria for annotations are the three

"C"s: correctness, completeness, and consistency. This review explores methods to enhance annotation quality through the use of advanced techniques that mitigate the limitations of single annotators. To support dataset creators, a standard operating procedure (SOP) is provided as supplemental material, outlining best practices for dataset development. Furthermore, the article underscores the importance of open datasets in driving innovation and enhancing reproducibility of DL research. By addressing the challenges and offering practical recommendations, this review aims to advance the creation of and availability to high-quality, large-scale datasets, ultimately contributing to the development of generalizable and robust DL models for pathology applications.

## Introduction

Deep learning (DL)-based automated image analysis (DL-AIA) is emerging as an important tool that promises to extract relevant information from microscopic images in an accurate, reproducible, and efficient manner. In fact, there is a substantial body of literature demonstrating the high capabilities of DL-AIA for various pathological tasks.[18,21,39,47,85,104,112] In these studies, supervised DL is a particularly popular approach, where a model is trained to recognize patterns based on input data (i.e., microscopic images) and map these to predefined output labels corresponding to the pattern of interest. Depending on the model architecture and type of output data, algorithmic predictions can range from image classification (i.e., classifying the entire image into categories, such as the tumor diagnosis for the case)[39] to object detection (i.e., predicting the location of objects of interest, such as mitotic figures, within an image)[18] to segmentation (i.e. classifying every pixel in the image to determine whether it belongs to an object of interest, such as the area of nuclei or tumor tissue)[47,Wilm, 2023 #4861] and thereby allowing for a wide range of pathological use cases.

The availability of high-quality, large-scale datasets, which comprise images along with their metadata and a collection of labels for each image or object of interest, are the precondition for development of supervised DL models and for testing their performance.[19,114] However, creating such datasets is challenging considering the high time investment needed for creation and at the same time the high risk for biases associated with image collection and label creation.[49,77,82,105] Insufficiently constructed datasets are recognized as a major source of errors in algorithms.[37,114] For example, a survey of toxicologic pathologists revealed that two of the three most common reasons why proof-of-principle studies of DL-AIA tools failed are related to the dataset due to (1) small dataset size and (2) insufficient dataset quality.[82]

This review article discusses all relevant steps of dataset creation for supervised DL and provides recommendations to ensure high dataset quality and quantity. A derived standard operating procedure (SOP) for dataset creators, covering all relevant steps, is provided as Supplemental Material. Another goal of this article is to summarize open datasets with veterinary samples that have been made publicly available to researchers to foster DL-AIA development.

For this article, we have defined different types of datasets and their subsets, as listed in Figure 1.[22] While some studies use alternative terminology,[92,99] the purposes of these datasets remain the same.[114] The training dataset provides the input data and output labels needed to update the DL model weights during training iterations. At regular intervals during the training, the validation dataset evaluates the model's progress on unseen images, guiding decisions to avoid underfitting (e.g. continue the training) or overfitting (e.g. stop the training). Model validation using the validation dataset should not be confused with clinical validation of a diagnostic test; therefore, some authors have used alternative terms for this dataset, such as "tuning set".[35,112] The term tuning set is, however, uncommon in the field of machine learning, and can likewise be misleading since also the model's parameters are tuned during training. Once the final model is selected on the validation set, its generalization performance is evaluated through statistical metrics and visual assessment on the test dataset, which is intended to be an independent representative of real-world data for the algorithm's application. Generalization performance measures whether the model is appropriate for predicting the patterns of interest on unseen data or whether they are overly specific to the development dataset. The development dataset, comprising the training and validation dataset, as well as the test dataset together form the primary dataset.

Whereas the subsets of the primary dataset are typically created within one process (and split after dataset creation), secondary test datasets are produced separately from the primary dataset. Thereby, secondary test datasets ensure avoiding spurious correlations of the images between development and testing (such as highly consistent color from the same staining batch), that do not sufficiently reflect the real-world variability. Annotations are needed for this dataset to allow calculation of statistical performance metrics.[49] The analysis dataset is composed of the images to which the final model will be applied to generate meaningful biomedical predictions in research or diagnostics. This dataset does not include annotation, but when used for research it includes tertiary metadata, e.g., patient outcome or data for clinicopathological correlations. The analysis dataset may include the same images of the primary and secondary dataset. While the analysis dataset is not necessarily created through the process discussed below, it defines the use case and thus determines the domains and real-world variability required within the other datasets, in particular the test datasets.[49]

Other authors distinguish between internal datasets (i.e. primary dataset derived from within a single organization) and external datasets (i.e. dataset derived from another organization), typically used in the context of secondary test datasets and performance evaluation.[81] In this review article, we avoid this terminology for two reasons: 1) it may be beneficial, depending on the intended use case of DL-AIA, to include data from multiple laboratories in the primary dataset,[49,114] and 2) external datasets are often annotated by a different group of experts, which may introduce label shifts compared to internal data, negatively impacting model development and testing. Instead, this review article discusses a similar concept for different image sources by using the term "domain". A domain is defined as the context of image creation that goes beyond the organization (e.g., specific image characteristics related to a specific laboratory, such as staining composition)[34] and includes aspects such as animal species, disease entities, sample preparation, and digitization devices/settings (see the next section for more details). Secondary test datasets may be "out-of-domain" (similar to the concept of external datasets) to evaluate the robustness of DL models under specific domain shifts.

Before discussing the three main elements of dataset creation – 1) images, 2) annotation software and 3) annotations – it is important to note that there is no one-size-fits-all approach to dataset creation. Dataset creators must make individual considerations and cost-benefit trade-offs based on (1) the pathology task (e.g., expected degree of error in human annotations), (2) the selected pattern recognition task (e.g., required level of annotations), (3) the intended use case (e.g., DL-AIA intended for entire whole-slide images or only "perfect" regions of interest), and (4) the available resources (i.e., laboratory resources and time availability of expert annotators). This article summarizes the various dataset creation methods applied in existing literature, reflecting the various project-specific requirements, but also makes recommendations for minimal standards. We hope that the subsequent section will provide readers with the knowledge to take these considerations into account when making decisions regarding dataset creation.

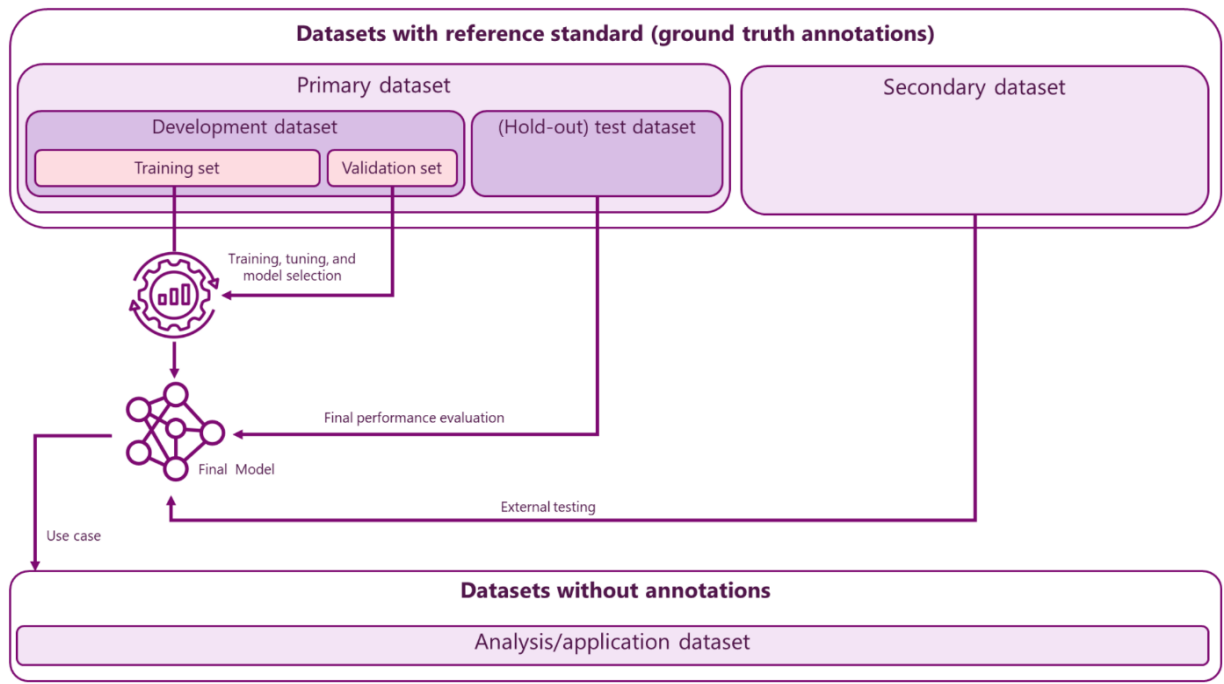

**Figure 1.** Comparison of different types of datasets and their subtypes along with their role in development, performance evaluation and application of deep learning models.

## Images

Whole-slide image (WSI) scanners are increasingly integrated into routine workflows of veterinary pathology laboratories and, therefore, WSIs are readily available for DL-AIA. As WSIs are the most common image type used for dataset creation in current literature, this article will focus on WSIs. Other digitization devices (such as cameras mounted on light microscopes or mobile phones attached to the ocular) and non-microscopy imaging modalities (such as gross images or electron microscopy images) are not specifically addressed in this review due to the paucity of research in this area.

Key considerations for image sets are the following:
- Sufficiently large number of images encompassing an adequate number of patterns of interest and background patterns, possibly stratified by subgroups
- Sufficiently large image variability, which is representative for the intended use case
- Inclusion of all relevant metadata for each case (e.g., patient/case information and image creation characteristics)

The selection of appropriate cases is critical, and clear inclusion and exclusion criteria should be defined that align with the intended use case of the DL-AIA. For example, inclusion of glass slides and/or WSIs from multiple laboratories may be considered to increase variability (see below).[11,12] Software for automated quality control processes, such as HistoQC, may help to identify unsuitable images with pronounced artifacts or other outlier features.[29,51] While these quality control tools are undoubtedly valuable for application datasets, caution should be exercised when removing cases with realistic artifacts and image variability from the primary dataset.

Image exclusion may be done when the image quality interferes with annotating the images with sufficient accuracy and images of poor tissue quality will also be excluded from the analysis dataset.[77] In fact, it may be of advantage to intentionally oversample certain artifacts to improve the models ability to distinguish them from the pattern of interest (see below).

Determining the total number of cases to include is challenging in advance, as there are no established methods to predict how many images are required to achieve the target algorithmic performance. The necessary dataset size depends on the complexity and morphological variability of the pattern of interest, as well as other structures present in the images. There are generally two strategies: 1) Maximal approach: Include as many cases as can feasibly be annotated within the available time and resources, aiming for the highest possible algorithmic performance.[19] 2) Iterative approach: Begin with a small dataset, then incrementally add more images as needed to reach the desired performance, typically focusing on those image parts that are problematic for preliminary DL models.

Although it may be convenient to include multiple samples from the same patient, such images are unlikely to introduce the same degree of variability as images from different patients, thereby providing less benefit for model development and testing. It also needs to be considered that images from the same patient, even if they originate from different tissue blocks or cytologic smears, can only be part of one data subset (training, validation or test dataset), due to the high similarity. Data leakage, i.e. distribution of data from the same patient across data subsets, would lead to overly optimistic performance evaluations [26] and must be avoided.

## Image variability and domain shift

Beyond the sheer number of cases, it is important that the images encompass the variability expected for the intended application, both for the pattern of interest (foreground) and for other background patterns present in the images.[114] It is well-known that DL models can learn biological and technical features specific to the image batch(es) of the training data (hidden variables).[10-12,40,49,50,57,63,114] Models often fail to predict the pattern of interest if the image features differ too greatly from those presented during training (covariate domain shift) This image variability is related to the multi-step process of image creation, encompassing tissue acquisition, tissue processing, slide preparation, image creation and image post-processing, often leading to a batch and laboratory specific image signature (Figure 2).[12,40,50,57,94,114]

Images from different species, including humans, often do not appear to cause major domain shifts, since tissue and cell morphology are often similar, at least for the pathology tasks with similar tissue and cell morphology between species.[8,46,70,71] However, caution is warranted for disease entities with species-specific features, for instance, the differing distribution of mammary carcinoma / breast cancer subtypes between dogs (with numerous mixed tumors) and humans, or the differences between red blood cell morphology between mammals and non-mammalian vertebrae with nucleated cells.

The included images should reflect the natural biological variability of tissue morphology.[49,63] This includes the need to represent all relevant disease entities, with potential oversampling of rare subtypes (class imbalance).[114] DL models usually do not generalize well across disease types (such as different tumor types) if they (or similar domains) were not included in training.[11,12,53] For example, a model trained to classify colon images as benign or malignant tumors showed a nearly 50% drop in performance when applied to the same task for breast and prostate tissue.[53] A mitotic figure object detection algorithm trained with soft tissue sarcoma images dropped in performance from an F1 score of 0.70 to 0.49 when applied to lymphosarcoma cases (images from same laboratory and scanner), whereas the in-domain performance for lymphosarcoma was F1 = 0.79.[12] In contrast, a DL model that was trained with multiple tumor types was able to detect mitotic figures with high performance in previously unseen tumor types.[12]

Nevertheless, even within the same images, variability exists. Consequently, it is an important consideration whether entire WSIs, individual tissue fragments or smaller regions within an image should be annotated, balancing time available for annotation and the expected degree of variability between image regions. For example, a tumor section may contain hundreds of thousands of nuclei, and it is neither feasible nor sensible to annotate all of them. However, the selected image regions should—depending on the intended application of the algorithm—not only include ideal (from the diagnostic standpoint) tumor regions but also more challenging regions, like necrotic or inflamed tissue, adjacent normal tissue, and image artifacts, to ensure adequate model generalizability. For mitotic figures, it has been shown that a model trained only on ideal (pathologist-selected) 10 high-power-field tumor regions can effectively detect mitotic figures in similar regions, but may perform poorly in different image areas (e.g., fatty tissue, thermal artifacts from cautery, or inking).[9,19] Thus, such an algorithm is not particularly suited to analyze entire WSIs and identify the ten high-power-field regions with the highest mitotic density.

Considering that there are everyday differences in slide preparation, particular attention should be given to producing samples across multiple batches (i.e., at different time points) to ensure realistic variability in tissue processing steps, such as section thickness and staining composition – unless the model is intended solely for use in a highly controlled laboratory environment. It should be ensured that cases from all relevant image subgroups are distributed across the batches. Alternatively, images from a WSI archive may be used, which naturally includes the real-world variability across the retrospective sample inclusion period. Application of color normalization in the training data should be avoided since it reduces the natural variability of images and can introduce additional, algorithmic artifacts.[55]

Different WSI scanners (see Patel et al. [84] for a summary of the different scanners) have been identified as a significant source of domain shift in studies that digitized slides with multiple scanners.[10,85,110] For example, a study on tumor segmentation models reported a decrease in the mean intersection over union (mIoU) of 0.38 when comparing test set performance of the in-domain scanner (mIoU = 0.82) with that of out-of-domain scanner (mIoU = 0.44).[110] Another study, evaluating a model that segmented lymph node tissue from the surrounding fatty tissue, demonstrated a significant drop in performance when trained on one scanner and tested on another, with performance decreasing by a factor of 4 (in-domain Matthews correlation coefficient, MCC = 0.81; out-of-domain MCC = 0.18).[56] While WSIs from different scanners exhibit obvious differences in color distribution (Supplemental Figure S1),[110] image post-processing methods intended to counteract this source of variability – such as extensive color augmentation (increasing variability in the training set) or image normalization (reducing color variability) – do not fully restore performance.[10,43] This indicates that WSIs from different scanners differ not only in color representation

(and image resolution) but also in other features, possibly related to hardware (such as optics) or scanning methods (e.g., depth of field, tile stitching, non-inclusion of non-tissue areas that contain artifacts such as dust).[84,110] There are clear recommendations to use multiple scanners for dataset creation, if the derived model is intended for a widespread application across multiple laboratories/studies.[10,114]

The degree of algorithmic bias (performance drop related to a domain shift) of the other potential sources of domain shift listed in Figure 2 has not, to our knowledge, been systematically evaluated. Whereas it is well-recognized that different laboratories produce different staining results,[34] it is not well understood if default augmentation methods during training can sufficiently overcome this laboratory variability. In human medicine, several studies on AI models for classifying HE images based on mutational patterns have demonstrated significant performance differences among images from different demographic/ethnic groups, related to underrepresentation of patients from certain demographic groups in training datasets (demographic bias).[100] Whether comparable differences exist among animal breeds remains unknown. Until further research reveals the impact of these unexplored variables, it may be advisable (depending on the intended application of the DL-AIA algorithm) to include these in datasets, and in particular in test subsets.

Figure 3 illustrates examples of false algorithmic predictions produced by a mitotic figure detection model when applied to arbitrary tumor regions of whole-slide images. The detector follows a standard anchor-free object detection paradigm based on the FCOS architecture [98] and was trained on the MIDOG++ dataset [12], which is, to our knowledge, the largest and most diverse resource for mitotic figure detection, spanning multiple tumor types, species, scanner vendors and laboratory origins. Despite this diversity, the dataset consists of ideal, pathologist-selected 10 high-power-field regions that are rich in viable tumors and contain only few artifacts. When deployed to less curated regions, the model becomes more sensitive to technical and biological artifacts that were previously unseen in training. As shown in Figure 3, dust particles on the slide (A-B), out-of-focus areas (C), sebaceous glands and other non-neoplastic structures (D-E), cautery artifacts (G-H), and surgical ink deposits (J-K) can all exhibit dark, hyperchromatic profiles that closely resemble the shape of mitotic figures. Similarly, cells in underrepresented tissue types such as apoptotic cells in embedded haemorrhagic areas (L), chondrocytes in cartilage (F), necrotic debris (M-N), superficial keratinocytes (O), and stain precipitate/tissue fragment (P) mimic texture and intensity patterns learned by the detector, especially when surrounding cellular morphology is degraded or missing. The principal causes of these false algorithmic predictions are 1) a domain shift between the clear training regions and the more heterogeneous data of the WSIs, 2) the presence of non-tumor or low-quality tissue that was underrepresented in the training data, and 3) genuine morphological ambiguity between mitotic figures and certain artifacts or cell death phenomena. This underscores the need for training datasets and model designs that explicitly account for such technical and biological variability when the model is intended for whole-slide inference.

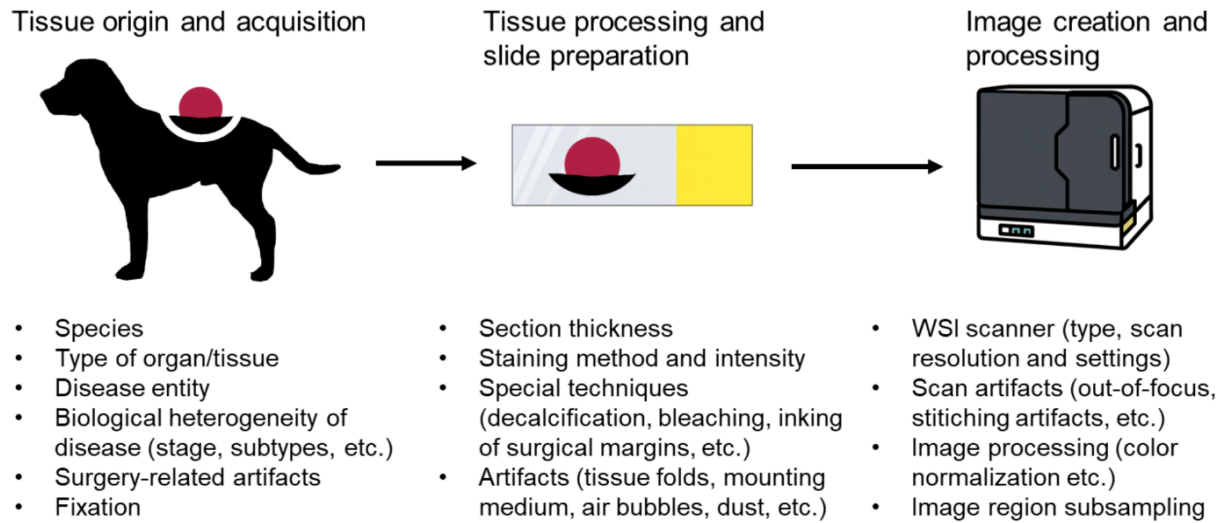

**Figure 2.** Sources of image variability that might impact the algorithmic performance, if not included in the training data, due to a domain shift. Dataset developers should consider these sources of image variability when selecting cases for the dataset based on their variability expected in the application dataset.

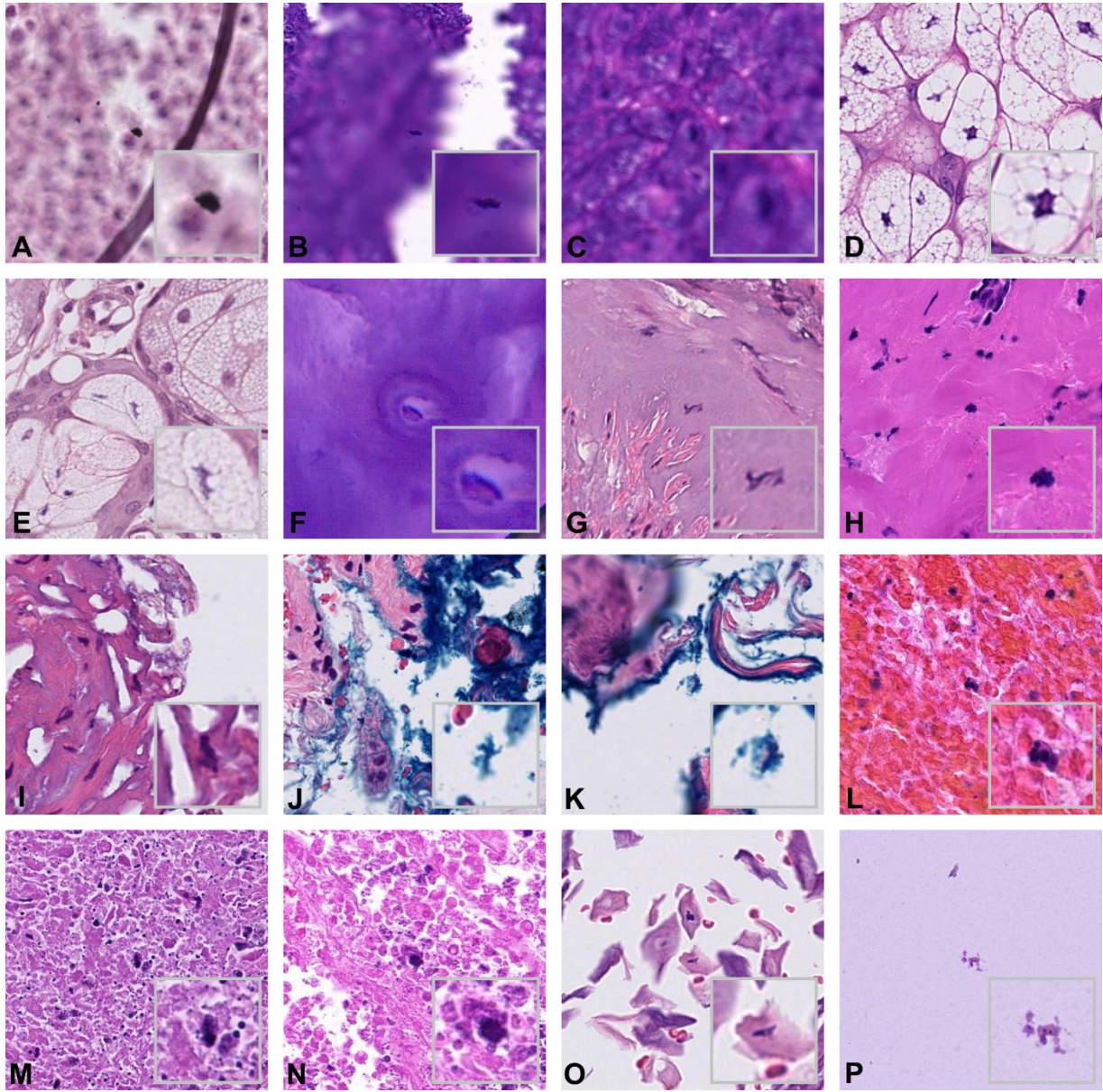

**Figure 3.** Examples of biological and technical variability of images between training and test datasets, that have caused false algorithmic predictions for detection of mitotic figures. Each panel shows the broader tissue context, with the model's false prediction highlighted in the corresponding zoomed-in inset. **A-B**) Dust. **C**) Out-of-focus. **D-E**) Sebaceous glands. **F**) Chondrocyte in cartilage. **G-I**) Cautery artifacts. **J-K**) Ink. **L**) Apoptopic cell within a haemorrhage. **M-N**) Necrosis. **O**) Keratinized epithelial cells. **P**) Stain precipitate or small tissue fragments outside of the tissue slice.

## Time efficient dataset creation

For most datasets, the most time-consuming step is the creation of annotations, particularly when object- or pixel-level labels are required, whereas the creation and curation of microscopic images is often comparably quick. However, there is an inherent association between the number of images and the time invested in annotations. Therefore, there are also specific image creation methods that can reduce the burden for creating annotations.

**Transferring annotations to further multi-scanner registered images:** For datasets that include scans of the same slide from multiple WSI scanners, image registration and subsequent transfer of annotations is a viable option.[38,72,110] The concept of multi-scanner-registered datasets is that each glass slide is digitized with multiple whole-slide scanners, but annotations are created only on a single reference-scanner image. These annotations are then spatially transferred to the corresponding images from other scanners, which enables realistic scanner-based augmentation of training data and enables cross-scanner performance evaluation. A prerequisite for this method is accurate registration between the reference-scanner images and the images from the additional scanners, ensuring that annotations map to the corresponding pixels. Because the same glass slide is scanned by all devices, the relevant differences are scanner-related and include (besides the intended factors of domain shift such as color distribution and depth of field) positional offsets, different resolutions and a 90° rotation. Unlike consecutive tissue sections, variations in the tissue (such as elastic deformation or tissue loss) or glass slide (air bubbles, etc.) should be the same between images (if scanned within a short period of time and properly cleaned before scanning), which simplifies the registration process.[36] Various open-source software for image registration is available,[36,72] but rigorous visual quality control is essential, and manual fine-adjustment may be needed for some images.[10] A challenge of this method is that cellular structures may be in different focal planes across scanners, leading to a loss of diagnostic detail in the out-of-focus image and requiring alignment of annotations, e.g., by a careful filtering scheme.[38]

**Minimizing the annotation effort by prioritizing most informative images (active learning):** Imagine having access to a large pool of unlabeled images but only a limited budget for expert annotation. Active learning can help determine which images should be prioritized. The main concept is to iteratively augment an initial dataset by selecting images or image regions that are expected to contribute most to model training (Supplemental Figure S2).[107] For WSIs, region-based active learning is typically used, in which only small informative regions within a WSI are proposed. This avoids annotating entire WSIs—an extremely time-consuming task—while ensuring that the regions most relevant for robust model development are included.[62,74,86,87] A comprehensive summary of active learning methods for medical imaging is provided by Wang et al.[107]; here we offer a brief overview. After training an initial DL model (using a small seed dataset), the unlabeled image pool is evaluated to estimate the informativeness each image or image region would provide if annotated and added to the dataset. Informativeness can be defined through model uncertainty or diversity in image representativeness.[86,107] Based on this informativeness metric, images or image regions are selected using a defined sampling strategy. The simplest strategy is to choose images in decreasing order of informativeness according to the available annotation budget, although more sophisticated strategies are recommended.[107] Of note, region size and the number of proposed regions per WSI/optimization circle may impact performance curves and effectiveness of an active learning pipeline strongly.[87] Once selected, the images are annotated (using the variable methods described below) and added to the dataset. This loop (i.e., selecting unlabeled images, annotating them, and updating the model) is repeated until the annotation budget is exhausted or the target performance level is reached. Prior studies have shown that active learning improves algorithmic performance for microscopic images compared to passive learning under equivalent annotation budgets.[62,86,87] While active learning can be highly valuable for reducing the time required to create a training dataset, it should not be used for test set construction to avoid biased image selection. Instead, an independent, application target-aligned test set should be constructed before running any active learning loop. Of note, active learning typically improves performance particularly for rare and boundary cases, and care may be taken to consider a suitable representation in this independent test set.

**Increasing image variability through synthetic data:** Synthetic images (also known as fake, simulated, or artificial images) are generated by computers in a way that they resemble real images.[83] Synthetic data is not produced from actual patient tissue samples, however, real data or mathematical models are used to guide the generation of synthetic images. Unlike conventional data-augmentation techniques (e.g., geometric or color transformations), which modify existing images during model training, synthetic images represent entirely new samples and can therefore provide higher-level diversity. Numerous methods exist for generating synthetic images, as summarized by Pantanowitz et al.,[83] with the most common approaches for microscopic images relying on generative adversarial networks (GANs) or diffusion models. An alternative to learning-based generative models was described by Mill et al.,[76] who developed a rendering-based approach that allows direct control over numerous image attributes (e.g., color distribution, cell morphology, connective tissue content, and artifact frequency). This fully parametric approach enables controlled, interpretable generation of images with any desired degree of variability (Supplemental Figure S3).

The primary benefit of synthetic data is the augmentation of training data without requiring slide production or annotation, as labels may (depending on the method used) be generated simultaneously along with the synthetic image. Once computational pipelines for image synthesis are established, large numbers of images can be produced at scale. Synthetic data can increase variability in the pattern(s) of interest [15] and/or transfer a new reference style to the existing images (e.g., simulating the appearance of a different scanner).[80] Useful applications include scenarios with limited access to real images (such as rare diseases or data-protection restrictions), substantial class imbalance, or limited annotation resources.[15,33,113] For example, Banerjee et al. [15] used conditional GANs to generate synthetic images for classifying normal versus atypical mitotic figures, motivated by the rarity of atypical mitotic figures (class imbalance). Another example is a public synthetic dataset of nuclei in human breast cancer comprising 20,000 image patches and 1,448,522 nuclei annotations—quantities that would be unrealistic for expert annotation.[33] These studies show that synthetic data can be beneficial when only moderately sized annotated datasets are available.[15,33,76,113] However, several limitations of synthetic images derived from generative models must be considered.[83,114] First, the variability of synthetic data depends on the diversity of the real data used to train or parameterize the generator; therefore, synthetic data is less valuable when based on a narrow or limited reference dataset. Generative models are known to suffer from mode collapse, i.e., only cover a part of the data distribution used for training.[27,58] Thus, synthetic data does not eliminate the need for at least a moderately-sized traditionally annotated dataset. Second, despite becoming increasingly realistic, synthetic images may contain subtle artifacts (such as checkerboard patterns, blurring, or excessive smoothing), that may be difficult for humans to detect but can bias DL models.[49,114] For this reason, test datasets must consist of real images,[49] whereas synthetic data may only supplement training sets. Furthermore, synthetic datasets should always be clearly labeled to avoid misuse.

## Annotation software

There is a variety of user-friendly annotation software, including both proprietary / commercial software (mostly combined with deep learning development frameworks; not listed in this article) and free open source software (Table 1).[7,14,17,44,60,61,64,67,75,106] Dataset creators need to choose software that supports viewing the image, allows creating annotations in the intended manner with high time efficiency (see next section), and generates datasets with a meaningful, reusable structure for AI development.[105] For proprietary software with integrated model development tools, it is important to ensure that it is possible (as it is for all open source software) to extract the dataset in a meaningful format to allow for long-term storage and reused in other projects. Most software are designed for a broad scope of annotation applications using microscopic (or even a wider range of biomedical) images,[7,17,44,60,64,67] while others have been developed for specific annotation tasks.[14,61] For example, SWAN enables swipe-based patch classification using a mobile device, allowing pathologists to annotate in offsite locations.[14] The annotation tool TissueWand is specialized for tissue wand polygon annotations using a mouse or pressure-sensitive stylus, i.e. a linear annotation is dragged in the center of the object and the annotations automatically spreads to the object borders.[61] PatchSorter focuses on two-dimensional annotation maps that clusters similar images patches in close proximity (generated with DL), which speeds up patch classification.[106]

Whereas most annotation software can open a variety of image types (including gross photographs and cellphone pictures through the microscope ocular), WSIs and z-stack WSIs are those with the highest requirements for viewing software. WSIs are not only extremely large files, but they also contain a pyramidal structure and most WSI vendors use their own proprietary file format. Most open source annotation software use open-source libraries, e.g., OpenSlide (https://openslide.org/), SlideIO (https://github.com/Booritas/slideio), or Bio-Formats (https://www.openmicroscopy.org/bio-formats/), which allow opening most WSI formats.[7,17,67,84] However, some vendors use proprietary formats that cannot be opened directly by vendor-independent software (e.g., Philips iSyntax) and therefore require conversion before use.[30,84] Based on one study, the file type is not expected to markedly impact algorithm performance.[53] However, there are ongoing efforts to harmonize WSI formats, such as the adoption of the DICOM format, which will facilitate standardization and interoperability across different software platforms and data exchange.[30,44] While DICOM is not yet widely utilized by dataset creators,[78] an advantage is that metadata is embedded within the WSI.

One of the main differences between annotation software lies in whether they are offline (mostly desktop-based) [7,17] or online (mostly web-based).[14,44,60,64,67,75,106]

Offline software is easier to set up for a standalone installation but offers limited collaboration capabilities (e.g., simultaneous annotations by multiple pathologists). Online platforms, on the other hand, share images by accessing a server, allowing collaboration with partners outside the organization while maintaining data privacy and protection through user management (e.g., secure user authentication, individual user rights and access), metadata encryption and prohibition of image download.

Depending on the pattern recognition task for the DL model, annotations need to be created at different levels. The software may allow different annotation shapes at the image level for image classification, at the object level (spot annotation, rectangle or circle at a fixed or variable size) for object detection and at the pixel level (polygons or tissue wand polygons) for segmentation. Time-efficient and accurate labeling should be facilitated through standard features (e.g., single-click annotation and keyboard shortcuts) and more advanced features, such as guided screening, blinded mode, and plugins.[7,67] The guided screening mode is particularly useful for large images (with many fields of view) to ensure complete annotations without skipping image regions. [7,67] In this mode, the large image is divided at a self-defined zoom level into patches of the size of the viewing field (with some percentage overlap), and the annotator is navigated through these patches sequentially. In the blinded mode other annotators do not see the label class of annotation, which may be for unbiased multi-expert annotations. Plugins are essential for visualization or image analysis and are often used for computer-assisted labeling or registration of two images (e.g., H&E and IHC of the same slide).[67,75,106] Much of the more recent annotation software focus on algorithmic support during annotations and thereby provide innovative tools to increase label efficiency (see annotations section).[67,75,106]

**Table 1.** Comparison of some relevant features of a selection of open-source annotation software.

| Feature | | QuPath [17] | SlideRunner [7] | Cytomine [64] | EXACT [67] | SWAN [14] |
|---|---|---|---|---|---|---|
| Application | WSI support | Yes | Yes | Yes | Yes | – |
| | DICOM support | Yes | Yes | Yes | Yes | – |
| | Offline / Online | Offline | Offline | Online | Online | Online, mobile device |
| Collaboration | User management | – | – | Yes | Yes | Yes |
| | Blinded mode | – | Yes | – | Yes | – |
| Annotation tools | Image classification | – | – | – | Yes | Yes |
| | Single click spot annotations | Yes | Yes | – | Yes | – |
| | Bounding box / Circle (variable size) | Yes | Yes | Yes | Yes | – |
| | Polygon | Yes | Yes | Yes | Yes | – |
| | Multi-user annotations | – | Yes | Yes | Yes | – |
| | Guided screening | – | Yes | – | Yes | – |
| | Plugins / Inference | Yes | Yes | Yes | Yes | – |

# Annotations

When annotating, a label is assigned to the image or objects of interest with an image and these labels are considered the ground truth (also referred to as reference standard), which serves as the desired output value during training and as the reference in the validation and test datasets for evaluating whether algorithmic predictions are correct or wrong. The ground truth represents the concept of truth generated by the most suitable method (gold standard method), balancing time investment with the following key quality criteria:

- Highest possible accuracy (i.e., the annotations truly represent the pattern of interest)
- Highest possible completeness/exhaustiveness (i.e. all patterns of interest are annotated in the images, which is particularly important for the test dataset)
- Highest possible consistency (i.e., the decision threshold between label classes and difficult background is the same throughout the entire dataset)

The gold standard method can be defined as interpretation by humans (manual annotations), humans supported by algorithmic suggestions (computer-assisted annotations), fully computerized annotations (see subsections below), or a superior method. For most pathology tasks, the gold standard is considered to be trained pathologists. However, human annotators are known to have well-recognized visual and cognitive biases, which can result in an imperfect ground truth.[1,32,88,89] For particularly challenging pathology tasks, human error can be significant enough to create a catch-22 situation: while a DL model is developed to predict the label classes with high accuracy, noisy ground truth data presented during training can lead to instability in model convergence. Likely much more problematic than training "confusion" is the presence of incorrect reference labels in the validation and test set, as these can lead to inaccurate evaluations of the model's performance, creating uncertainty and making it difficult to interpret performance metrics.[52] Systematic error in the annotations will lead to discrepancy with interpretations by other pathologists and bias inference results on applications datasets. It is important to note that a true (error-free) ground truth does not exist for most pathology tasks, however, we recommend placing emphasis on high quality of labels (in particular for the test datasets), which can be achieved by a well-planned annotation workflow and numerous advanced annotation methods (Table 2). The few superior gold standard methods often evaluate the pattern of interest from an independent level, such as patient outcomes (patterns of interest are undefined morphological malignancy criteria) or mutational patterns (determined by molecular methods), which models can learn to correlate with corresponding histological patterns.[31,85] These superior ground truth methods are not further discussed in this review, due to their paucity in veterinary datasets.

Depending on the intended pattern recognition task for the DL model, different annotation shapes need to be applied (including global image labels, spot annotations, bounding boxes, polygons/masks), with higher level of label details (granularity) typically requiring more time investment (Figure 4).[105] The most suitable annotation shapes should be determined based on their time investment, balancing the richness of information most beneficial for the pattern recognition task with the number of annotations that can be created within the same time frame.[95] When time efficiency is a critical concern, some projects may opt for sparse annotations, which involve intentionally annotating only a limited percentage of the pattern within the selected image.[114] This lack of completeness in annotating all patterns introduces limitations for DL model development and is therefore typically combined with fully annotated images.[114] It is important to note that sparse annotations are not suitable for validation and test datasets.

For image classification, there must always be at least two label classes. For object detection and segmentation, annotations may have only a singular label class (e.g., tumor tissue)[5] or multiple classes (e.g., number of argyrophilic nuclear organizer regions, AgNORs, per tumor nucleus with ten label classes)[41], whereas all non-annotated objects or pixels belong to the background class. For some pathology tasks, it may be useful to additionally annotate particularly challenging (for the DL model) background patterns as "hard negatives".[19,63] The primary benefit of hard negatives is that they can be shown to the DL model during training at an overproportionate ratio to reinforce the model's ability to distinguish patterns with overlapping morphology. These hard negatives include 1) objects with an overlapping morphology to the pattern of interest (e.g., apoptotic cells for a mitotic figure dataset), for which pathologists may also struggle to classify them correctly, and 2) pattern that may be difficult for the DL model to learn due to their rarity in some datasets, but are obvious for pathologists (e.g., green ink with a shape of chromosomes in metaphase could be confused with a mitotic figure, Figure 3J and K). A secondary benefit of hard negatives is that they can be used when forming a majority vote by multiple annotators (see below).

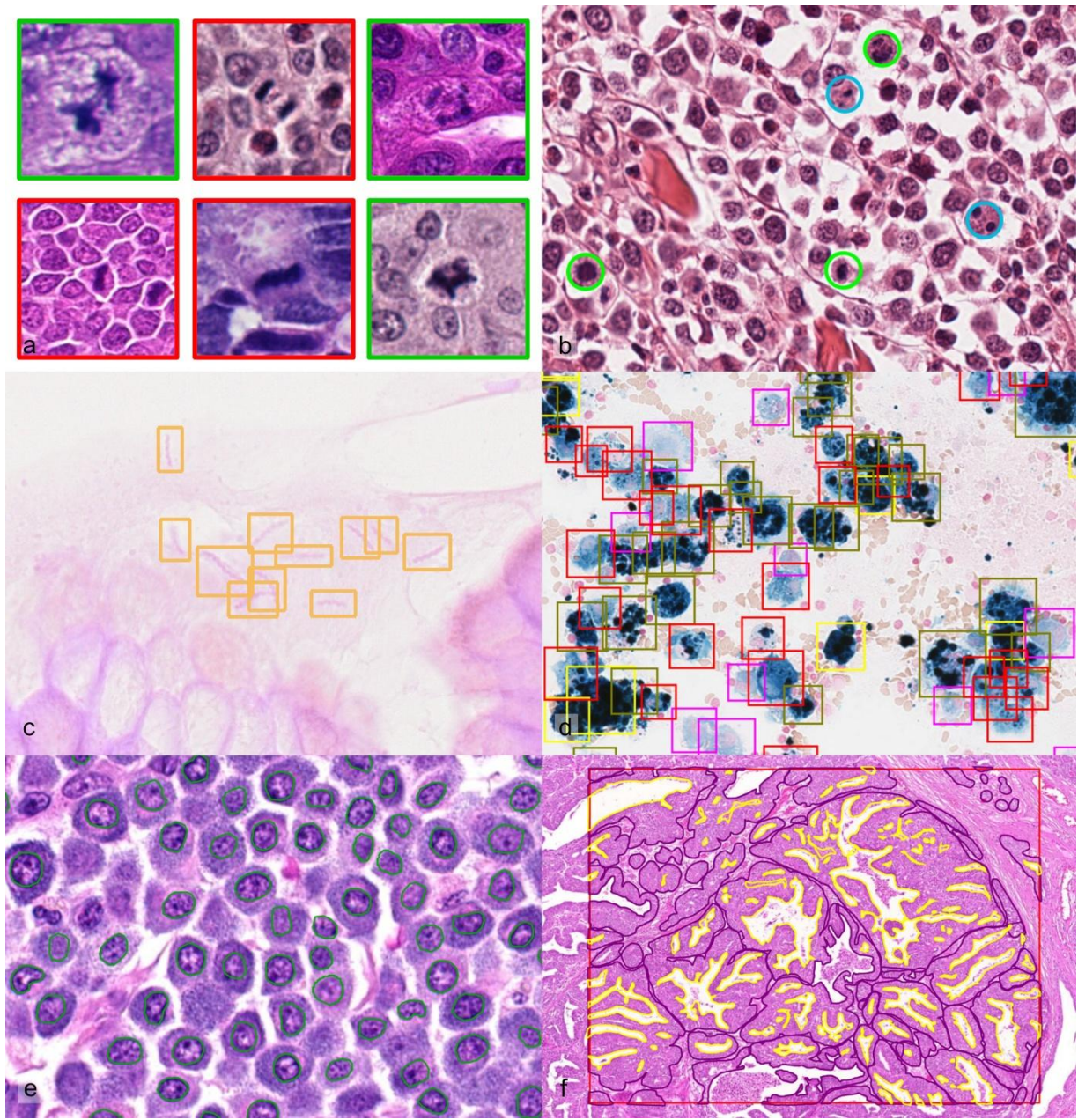

**Figure 4.** Comparison of different annotation shapes with increasing granularity. **a)** Classification labels of image patches with labels for normal (red) and atypical mitotic figures (green). Images of different canine and human tumors from an open dataset.[16] The difference in stain color between the patches is related to slides being stained at different laboratories and scanned with different WSI scanners. **b)** Single click spot annotation in the center of mitotic figures (green) and hard negatives (blue). For better detectability the annotation is displayed as a fixed size circle. Image of a canine cutaneous mast cell tumor from an open dataset.[19] **c)** Bounding box annotation with variable size, labelling helicobacter spp. in a canine gastric biopsy. **d)**

Bounding box annotations with variable size, labelling 5 different classes of hemosiderophages. Image of a bronchoalveolar lavage of a horse with exercise-induced pulmonary hemorrhage from an open dataset.[70] **e)** Polygon annotation delineating the contours of neoplastic nuclei. Image of a canine cutaneous mast cell tumor from an open dataset.[47] **f)** Region of interest annotation (red box) and polygon annotations delineating the contours of neoplastic epithelium (purple is inclusion area and yellow is exclusion area). Image of a canine mammary carcinoma.

**Table 2.** Comparison of some annotation methods that can be used to generate labels for microscopic datasets.

|  | Manual | Computer-assisted | | Stain-registered transfer of computerized annotations |
|---|---|---|---|---|
|  |  | **Previously labeled images** | **Unlabeled images** |  |
| Purpose | Simple setup; initial dataset for algorithmic-augmented labeling | Improve quality | Increase label speed | Eliminate manual annotation workload |
| Applied method | Pathologists' decisions | Detection of missed candidates or annotation maps with pathologists' decision | Expert-Algorithm-Collaboration | Create HE and IHC image from same section; computerized detection of IHC signal, extraction as annotations and transfer to HE image |
| Required pathology skills | High | High | High | Low |
| Required computer science skills | Low | High | High | High |
| Limitations | Visual and cognitive bias of humans; high time investment | algorithmic error propagation; confirmation bias of humans | algorithmic error propagation; confirmation bias of humans | Weak staining and background staining of IHC, loss of tissue during re-staining |
| Options to improve | Clear decision criteria; multi-annotator majority vote; image registration with another staining/IHC, etc. | Highly sensitive predictions (include hard negatives to reduce confirmation bias), Highly diligent review of predictions by expert | Highly specific predictions? highly diligent review of predictions by expert | Improve laboratory workflow |

## Manual annotations

The default annotation method for microscopic datasets are manual annotations by a single trained pathologists.[78] While the advantage of this approach is the perceived simplicity, humans are prone to numerous cognitive and visual biases that can impact the quality of annotations.[1,32] While a single annotator may be appropriate for simple pathology tasks (e.g., delineating neoplastic nuclei), the degree of annotation bias increases with more complex tasks (such as annotating mitotic figures). Several studies have shown that a single-rater dataset can contain a significant degree of error, and that reduction of label errors through more sophisticated annotation methods can markedly improve performance of the derived DL model.[45,109] One particularly important type of error in microscopic images is low inter-rater consistency between annotators.[20,41,42,109,112] For example, the number of mitotic figure annotations by 12 pathologists in 50 regions of canine cutaneous mast cell tumors ranged from 1,324 to 4,412, differing by a factor of up to 3.3.[109] When comparing these annotations with the majority vote of the 12 pathologists, the performance of each annotator ranged between an F1 score of 0.68 to 0.86 (where a score of 1.0 indicates perfect agreement to the consensus). Another study revealed the performance of 6 annotators on the test subset of a dataset for multinucleated tumor cells, with the annotators F1-scores ranging from 0.316 to 0.622 when compared to a computer-assisted ground truth created by a separate pathologist.[20] This high degree of inter-rater disagreement for certain tasks should be carefully considered when creating datasets with a single annotator, especially if the derived DL model is intended to be applied by different annotators. DL models are likely to assimilate to the annotators decision thresholds and there might be an inevitable difference between the DL models predictions (even if the model was perfect compared to the single-rater ground truth) and the interpretation of other pathologists.

Methods to improve label quality of manual annotations include:

- Annotator experience: A high level of experience fosters accuracy and consistency in decision-making (see below).
- Detailed annotation instructions: Providing clear labeling instructions and label class definitions, along with example images, can improve label quality.[77,105] It should be emphasized that omitting difficult patterns due to decision-making challenges is not a viable option for dataset creation.
- Pilot studies: Conducting a pilot study, where the same slides are annotated by multiple annotators and/or by the same annotator after a washout period, can help to assess inter- and intra-rater agreement. These metrics serve as indicators of label quality. Based on the results, annotation methods and instructions can be optimized. Providing feedback to annotators on their performance may raise awareness of the importance of diligence and adherence to instructions.
- Training phase: A training phase prior to annotation is recommended to familiarize annotators with the annotation software and annotation task.
- Majority vote: Using a majority vote from multiple annotators can harmonize individual sensitivity and specificity trade-offs, resulting in labels that reflect the average pathologist's interpretation (see below).[18,21,47]
- Reevaluation: Reevaluating annotations by the same annotator(s) can reduce label inconsistencies and correct accidental errors. Inexperienced annotators may exhibit a learning curve during dataset creation, leading to

- inconsistencies between earlier and later annotations. Accidental errors may also arise due to time pressure and fatigue.
- Cross-checking: Cross-checking annotations with an experienced pathologist can help identify misinterpretation of the labeling instructions and errors in subsequent annotations can be avoided.[47,78]
- Special staining methods: Registration with specific staining methods (e.g., histochemical stains or immunohistochemistry) can support annotators to better classify difficult patterns (see below).

The required level of expertise of annotators depends on the specific annotation task. For example, annotations of neoplastic nuclei may be performed by veterinary students,[47] while labeling mitotic figures requires a higher degree of expertise.[23,109] Highly experienced annotators are generally expected to produce fewer label errors; however, this may not hold true for tasks requiring minimal medical background, such as interpreting the intensity of special stains.[66] While some experienced pathologists may have limited time availability due to other commitments, less experienced annotators, such as veterinary or PhD students, may be able to dedicate more time and effort to the project, which may positively impact label quality. For less experienced annotators, prior training and regular cross-checking by experienced pathologists can improve label quality.

To avoid placing extensive workload onto one annotator, some groups split the images among multiple annotators.[4] However, different annotators may have different decision thresholds resulting in a label shift between annotators and label inconsistency over all images. In such cases, minimizing inter-rater variability is critical, for example through clear annotation instructions. It is advisable to have all annotators label the same images of a small subset to allow identification of inter-rater variability between the annotators, which can then be accounted for in subsequent DL development steps. For certain image-level labels, such as case diagnoses, annotations can sometimes be retrieved from existing medical records, reducing the need for additional annotators but still requiring careful quality control.[65]

For majority voting, an uneven number of annotators (typically three) is necessary, along with a blinded review process.[45,109] Studies have shown that majority voting improves label quality, resulting in more accurate labels that reflect the average pathologist's interpretation by balancing outlier opinions.[45,109] An important consideration is the number of annotators required for each label, balancing the overall label effort with label quality. For object- and pixel-level annotations, two approaches can be used: 1) One annotator screens the slides, and patches of the annotations are reviewed by a second annotator. Discrepancies are resolved by a third tie-breaker annotator. 2) Two annotators independently screen the images, and all discrepant labels or objects with only one annotation are reviewed by a third tie-breaker annotator. The first approach is more time-efficient, while the second promises greater label exhaustiveness. For annotations tasks involving a single label class (e.g., mitotic figures in tumors), a hard negative group should be included alongside the pattern of interest annotations in a blinded manner to avoid confirmation bias.[23]

For tasks requiring annotation of difficult patterns in HE images, additional information provided by special stains (e.g., histochemical stains or immunohistochemistry, IHC) can support the annotator's decision-making process.[11,42,71] For this method following laboratory steps are needed: 1) apply first stain, 2) scan, 3) destain, 4) apply second stain, 5) scan, 6) image registration (see example of *Helicobacter* spp. in Supplemental Figure S4). While this approach can improve label efficiency (by not overlooking objects and by removing the need for a majority vote) and may improve on the key quality criteria, it has limitations. These include costs associated with additional staining, potential imperfections in the second staining (e.g., tissue loss during restaining, staining residue, etc.), and an increased sensitivity of the special stain compared to the identifiability of the HE morphology, which may lead to label shifts.[42]

To facilitate the annotation work, annotators should be provided with appropriate annotation software (see above) and computer hardware, such as a computer mouse or stylus and a suitable monitor.[77,78] While the impact of hardware on annotation accuracy has not been thoroughly evaluated, it certainly affects label efficiency and work ergonomics.[28] For example, depending on individual preferences, using a stylus may speed up polygon annotations.

## Computer-assisted annotations

Computer-assisted annotation methods (also known as human-algorithm collaboration) are commonly used for dataset creation in both human and veterinary research.[78,82] The key feature of this approach is the collaboration between a computer algorithm (not limited to deep learning models) and an annotator, where the algorithm generates predictions on dataset images that are subsequently reviewed and refined by the annotator. While computer-assisted annotation methods can significantly improve efficiency or label quality, dataset creators must be aware of potential biases, including algorithmic errors and biases introduced by human-algorithm interaction. Particular care must be taken to avoid introducing bias into the test dataset. The prerequisite for these tools is the availability of a sufficiently high-performing algorithm, which can be either task-generic (e.g., a segmentation model for any round structure)[28] or task-specific (e.g., detection of mitotic figures in mammary carcinoma).[8] Task-specific algorithms often require an initial manually annotated dataset for its development.

Based on our literature review, computer-assisted annotation tools can be grouped into three categories, each with distinct applications and benefits:

- Algorithmic inference on unlabeled images with subsequent annotator review. These tools aim to increase labeling speed.
- Real-time algorithmic modification during creation of manual annotations. These tools aim to enhance annotation granularity while maintaining high efficiency.
- Algorithmic inference on previously labeled images and subsequent annotator review. These tools aim to increase label quality (completeness or consistency).

Algorithmic inference on unlabeled images uses algorithms to predict new labels, reducing the amount of manual interaction required from annotators.[70] The algorithm generates suggestions, which the annotator reviews and, if necessary, corrects by removing, adding, or modifying the size/shape of the annotations. Studies have demonstrated that this method can significantly reduce the time investment for labeling and, when high performance algorithms are used, even improve annotation quality.[66,69] However, the most relevant source of error in this approach is confirmation bias, where annotators may accept incorrect algorithmic predictions while dismissing contradictory information.[89] Previous studies have shown that annotators may fail to identify errors such as missing annotations, non-maximum suppression artifacts, incorrect label classes, and false-positive predictions.[45,66,69] Therefore, high diligence during expert review is essential to mitigate these risks. Algorithms with a balanced or specific detection threshold may be particularly effective in addressing the objective of increasing label efficiency.

Real-time algorithm-annotator interactions are often designed to create highly granular polygon annotations with minimal effort.[3,28,61,95] Annotators provide low-granularity input, such as a mouse click, a line inside the object, or a bounding box around the object, and the algorithm automatically refines the annotation to outline the object at the pixel level. These tools are known for their high efficiency and typically have a high accuracy in generating polygon annotations.[28,61] However, human review and corrections are necessary to ensure high label quality.

After the initial manual dataset is completed, further computer-assisted methods can be applied to improve annotation quality. These methods typically involve training deep learning models on the initial dataset to either: 1) detect missed candidates [8,19,23] or to 2) remove annotation errors.[8,67,70] Small and rare objects (such as mitotic figures) are often overlooked due to the complexity of microscopic images combined with time pressure or annotator fatigue. To improve label completeness, deep learning models with high sensitivity can be used to detect missed candidates. These models are designed to include nearly all missed candidates, at the cost of generating false positives.[8,19,23] The high number of false positives has a positive effect, as it reduces confirmation bias during subsequent annotator review.

Annotation errors can be detected using annotation maps, which visualize patches of annotated objects based on a predicted relationship.[8,67,70] These maps group patches with high similarity in close proximity and display the previously assigned label class. Annotators can explore these maps to identify and correct errors in label classes. For example, patches of the same label class located at opposite ends of the map are likely to represent obvious label errors that can be easily corrected. Conversely, regions where two label classes blend into each other often represent borderline morphologies that are inherently difficult to annotate. Correcting these labels may introduce bias, so care must be taken during review. An example of annotation maps was demonstrated in a dataset of hemosiderophages in equine bronchoalveolar lavage fluid.[70] In this study, hemosiderophages stained with iron stains were classified into five categories based on the degree of intracytoplasmic hemosiderin (blue pigment). These categories represent continuous degrees with artificial thresholds, which can be challenging to apply consistently. Annotation maps, which ordered patches based on a regression score, were used to verify the consistency of the applied thresholds and thereby improved label quality.

## Computerized annotations

There are few annotation methods used to increase the number of labels that do not rely on annotator interaction at any point, making them fully computerized. These time-efficient labeling methods include pseudolabels and stain-registered transfer of computerized annotations.

Pseudolabels are annotations derived through algorithmic inference on unlabeled images without verification by an annotator. For this approach, an initial, manually labeled dataset must be available to train sufficiently accurate algorithms. However, this creates a self-reinforcing problem: an initial small dataset may lead to a biased algorithm and subsequently generate extensively flawed pseudolabels (resulting in error propagation), thus only slight advantages for model training can be expected.[2] On the other hand, already highly accurate DL models (based on large initial dataset) may benefit little from training with pseudolabels. Some studies generate pseudolabels for high confidence predictions while requesting expert review for low-confidence predictions and thereby aiming at reducing annotation effort and ensuring label quality.[59,62] Importantly, pseudolabels should never be used for testing datasets.

Stain-registered transfer of computerized annotations is achieved by creating HE images and special stains (including histochemical stains and IHC) for the same tissue sections. The special stain is used to generate the ground truth, which is subsequently transferred to the HE image (Supplemental Figure S5). Annotations can be automatically generated with high label quality from special stains by creating a binary mask through simple algorithms, including color deconvolution followed by filters and thresholding.[5,13,24,54] The transfer of annotations enables deep learning (DL) models to be trained to identify patterns of interest in the corresponding, morphologically more complex HE images. Manual annotations are not required for this approach,[5] although some authors have used manual refinement of the automated labels.[24] The advantage of this approach is that a large dataset, including complete annotations of entire WSIs, can be generated with no or minimal annotation labor. Given the low degree of error associated with this approach, it is potentially also suitable for the test dataset, provided quality control of the automatic annotations is conducted. Nevertheless, the use of a second test dataset with human-curated annotations may be advisable to confirm the high accuracy of the automated labeling method.

This special stain-based annotation approach has been utilized in current literature for the creation of WSI segmentation masks of neoplastic mammary epithelium in dogs (using cytokeratin IHC as the special stain),[5] colonic epithelium in humans (using EPCAM IHC),[54] and collagen stroma in canine tumors (using Hematoxylin-Eosin-Saffron stain).[13] Mehrabian et al. [73] employed this approach for the segmentation of epithelial lung tumors (using cytokeratin immunofluorescence) exclusively for pretraining a model, which was subsequently fine-tuned with human annotations, reducing the annotation time investment by 70%. Bulton et al. [24] applied a slightly different approach by generating segmentation masks of neoplastic prostate epithelium from IHC images and subsequently trained a DL model for the IHC images. The DL models then generated segmentation masks (pseudolabels) for additional IHC images, which were only then transferred to registered HE images.

## The evolution of dataset creation for mitotic figures

This section aims to demonstrate the challenges of dataset creation using the example of datasets for mitotic figures (both human and veterinary samples). We compare relevant annotation methods and explore how these methods impact the derived DL object detection models. Mitotic figures represent a particularly challenging pattern for both pathologists and computer vision systems, necessitating especially high standards for ensuring dataset quality. From the six open access datasets listed in Table 3, it is evident that the applied methods for dataset creation have become increasingly complex over time to address challenges with annotation bias.[8,12,19,23,90,91,101,102] Based on the number of annotators and the use of computer assistance, we have classified these annotation methods into three versions. Another version of dataset creation, currently being investigated and discussed below, is the use of IHC-assisted methods (version 4).

One of the earliest mitotic figure datasets used a ground truth derived from a single annotator (version 1; e.g. the MITOS dataset).[91] This dataset was used for a computer scientist challenge, where other research teams competed to develop the best-performing algorithm. A high F1 score of 0.782 was archived by the winning team, however, this high performance can be largely attributed to data leakage (i.e. image patches from each case were included in both the training and test datasets). The primary source of label bias is the correctness of the mitotic figure annotations. Previous studies have shown that pathologists apply variable thresholds when classifying mitotic figures against the background (e.g., apoptotic cells), which can result in differences in the number of annotations of up to a factor of 3.5 when analyzing the same images.[18,109] While it is typically unknown whether the annotator applied sensitive or specific thresholds, the individual annotator has a great influence on the derived DL model.

Version 2 datasets attempted to address inter-annotator differences by employing a multi-expert majority vote for formulating the ground truth (e.g. the AMIDA 13, MITOS-ATYPIA and TUPAC datasets).[90,101,102] The AMIDA13 dataset is a good example of this version and it provides insight into the agreement between their annotators.[102] Two annotators independently screened the 23 images, resulting in 1,088 annotations by annotator 1 and 1,599 annotations by annotator 2. Of these, 649 objects were annotated by both annotators (31.8%; accepted for the final dataset), while 1,389 objects (68.2%) were labeled only by one annotator. The latter annotations were evaluated by additional pathologists who agreed that 434 of the 1,389 objects (31.2%) were consistent with mitotic figures. This multi-expert majority vote aims to balance the individual pathologists' thresholds towards an average pathologist's interpretation. Two studies have evaluated the number of annotators needed for a consensus and both concluded that three expert annotators provide an ideal balance between increased label accuracy (defined as the average pathologists' interpretation) and additional time investment. [45,109] While this mean decision threshold is considered beneficial for achieving acceptance of the DL model by the majority of pathologists, it should be acknowledged that there will always be pathologists with divergent thresholds, meaning this approach does not completely resolve the issue of label correctness. Even the same annotators will not result in the same ground truth, revealing that there are challenges to apply the decision thresholds consistently. For example, when the ground truth objects of the AMIDA13 test set were reevaluated by the same annotators, only 379/533 objects (71.1%) were considered to be true mitotic figures upon reevaluation.[102] Another relevant source of label bias in this version 2 method is the completeness of annotations. This is also supported by the data from the AMIDA13 computer scientist challenge, for which all "false positives" predictions by the best performing DL models (i.e. predictions not marked as ground truth labels) were reevaluated. It was found that 61/208 (29.3%) "false positives" were indeed consistent with mitotic figures and had been missed by the initial annotation workflow.[102]

Successive datasets incorporated computer assistance for identification of missed candidates (version 3; e.g. the MITOS_ccMCT, TUPAC alternative, MITOS_CMC, and MIDOG++ datasets) .[8,12,19,23] This approach is applied after all images have been annotated using a version 2 workflow (Figure 5), with an emphasis on reducing annotator bias by predicting candidates at high sensitivity (i.e., intentionally including many false positives). Previous studies have demonstrated that the measured test set performance of the DL model (using the same images and training methods) can be improved from F1 = 0.755 to 0.820 [19] and from F1 = 0.707 to 0.785.[8]

Given that this annotation workflow of version 3 has become quite time-consuming, alternative methods have been evaluated that use re-staining of the same tissue sections with IHC of pHH3 (version 4),[11,18,42,45,97] a marker that highlights mitotic figures throughout prophase (not consistently recognizable in HE images) to anaphase. Tellez et al. [97] have automatically detected the brown signal of mitotic figures (color deconvolution) and used a DL model (trained on 2,000 manually annotated objects) to distinguish staining artifacts from real signals (pseudolabels for a relatively easy computer vision task). These pseudolabels were then transferred to the registered HE images, enabling the creation of a large-scale WSI dataset with only two hours of annotation work. However, the derived model did not score particularly high performance, despite the large amount of training data. Based on our experience, many pHH3-positive structures cannot be recognized in HE images based on morphology (e.g., early mitotic figures, overstained cells, out-of-focus image regions, or marginally sectioned cells).[11,42] This may have introduced label noise in the annotation method by Tellez et al. [97] and impaired model training. A subsequent study by Ganz et al. [42] used the pHH3 slide as decision support for pathologists during annotations in HE images. This approach aimed to improve label completeness (eliminating the need for computer-assisted missed candidate screening) and enhance annotator consistency by helping pathologists better distinguish mitotic figures from imposters. While datasets from single annotators showed much higher agreement, this annotation method did not result in improved model performance. This outcome was attributed likewise to an information mismatch between HE and pHH3, which led to a label shift by annotators (i.e., increasingly annotating morphologically less certain mitotic figures). Further improvements of this annotation approach, such as by majority voting (blinded to pHH3) for borderline morphologies, was suggested.

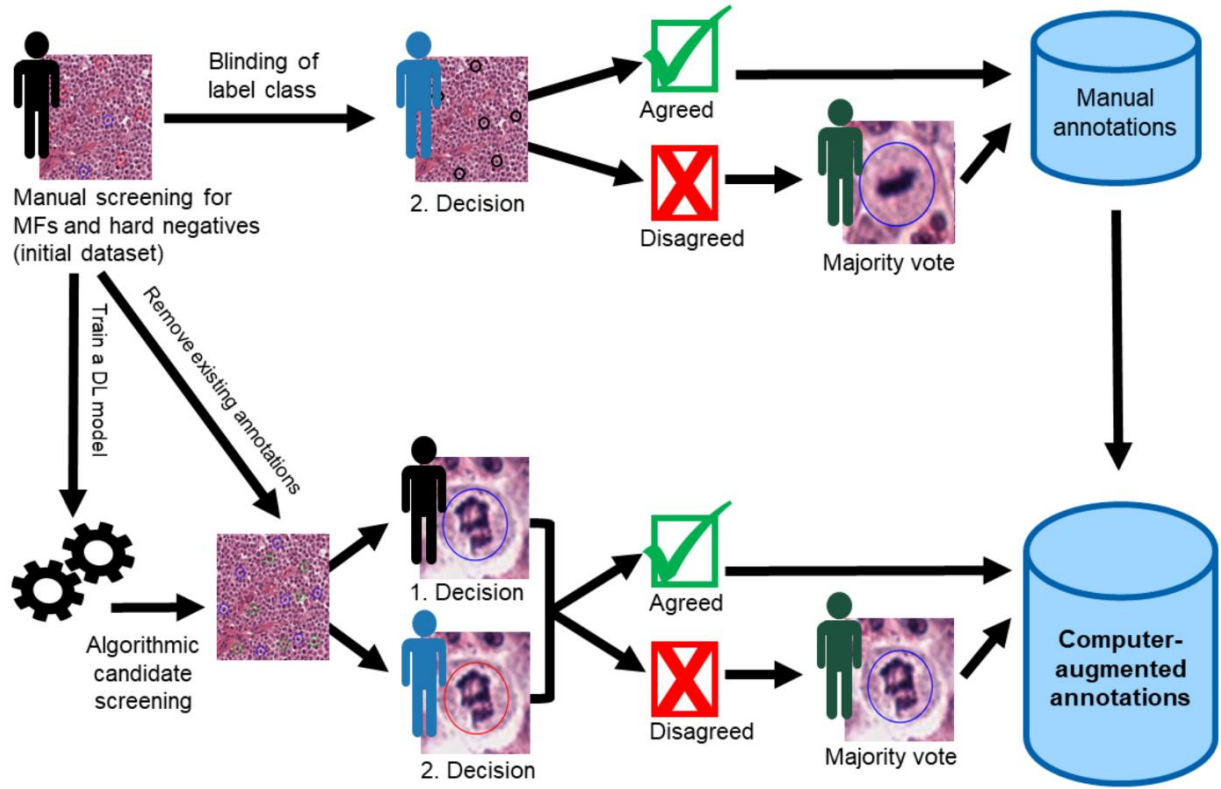

**Figure 5.** Annotation workflow for a version 3 dataset. First an initial dataset is created by a pathologist screening all images for mitotic figures and hard negatives (background structures resembling mitotic figures). The initial dataset is 1) blinded for the label classes and presented to further pathologists for majority voting and 2) used to train a deep learning model that is used to screen for mitotic figure candidates that were missed by initial screening. The model predicts candidates with a high sensitivity, resulting in an intentionally high number of false positives, which removes a confirmation bias in subsequent majority voting.

This graph was modified from Bertram et al. [23]

**Table 3.** Comparison of creation methods for mitotic figures datasets using human and animal tumors images, ordered by annotation method version and year of publication.

| Dataset acronym or reference | Year | Annotation method | | | |
|---|---|---|---|---|---|
| | | Annotators | Computer-assisted missed candidate search | Other | Version |
| MITOS [91] | 2012 | 1 | – | – | 1 |
| AMIDA 13 [102] | 2013 | 4 (majority vote) | – | – | 2 |
| MITOS-ATYPIA [90] | 2014 | 3 (majority vote) | – | – | 2 |
| TUPAC [101] | 2016 | 4 (majority vote) | – | – | 2 |
| MITOS_ccMCT [19] | 2019 | 2 (consensus) | Yes | – | 3 |
| TUPAC alternative [23] | 2020 | 2 (consensus) | Yes | – | 3 |
| MITOS_CMC [8] | 2020 | 3 (majority vote) | Yes | Annotation map | 3 |
| MIDOG++ [12] | 2023 | 3 (majority vote) | Yes | – | 3 |
| Tellez et al. [97] | 2018 | 0 | – | IHC-based pseudolabels and transfer to HE images | 4 |
| Ganz et al. [42] | 2024 | 1 | – | IHC-assisted annotations in HE images | 4 |

## Further considerations

Before using the dataset for model development, it is recommended to conduct data cleaning (which involves detecting and correcting or removing errors, incompleteness, and inconsistencies) and quality control of annotations and images.[105] For example, missing metadata can be added and obvious errors, such as duplicate annotations and wrong annotation types, can be automatically removed. The development of an initial deep learning (DL) model may also provide insights into the dataset's quality and reveal potential areas for improvement through augmentation of the dataset.

Effective dataset management requires thoughtful storage, backup, and long-term accessibility. Storage should match access needs: fast local or network solutions during annotation and model training, and slower archival systems after project completion. To prevent data loss, at least one independent backup should be maintained on a separate medium (e.g., cloud, external drive, server, or publish data repository). During active dataset creation, backups should be performed regularly, ideally with version control to track changes. Equally important is the choice of image and annotation formats. Proprietary or third-party formats may become unreadable if their software is discontinued and open, standardized solutions should be preferred. In digital pathology, DICOM is a proposed solution for a well-documented, vendor-independent standard for WSI and annotations, that ensures interoperability and long-term accessibility.[30]

Proper documentation of all dataset creation steps, as well as a detailed description of the distribution of the final dataset (e.g., the number of images/cases and annotations as well as their distribution across image domains and disease subgroups), is essential for ensuring transparency regarding potential dataset bias and enabling informed decision in subsequent DL model development steps.[22,35,49] For example, metadata on patient subgroups will allow a stratified performance evaluation possibly revealing sources of algorithmic error (hidden stratification).[49] However, it has been shown that public datasets with human samples often lack relevant information,[96] highlighting the need for particular attention to this matter. To support authors in complete reporting of dataset characteristics, Elfer et al.[35] have published a reporting guideline for annotations.

Although splitting the primary dataset is typically done after dataset creation and before DL model development, it is based on dataset characteristics. Splitting into three subsets (training, validation, and test) must be done at the patient level to avoid data leakage between subsets, for which information on the patient ID for each image is needed.[49,114] While the test dataset should be representative of real-world applications, a random patient-wise splitting approach may not be ideal for small datasets. In such cases, stratified splitting—where groups are defined based on, for example, image domains or the density of annotations per image —is often preferred. These dataset characteristics (patient ID, subgroup categorization) should be encoded in the final dataset. The dataset splitting scheme should also be documented so it can be used consistently across studies enabling comparison of the results.

Dataset owners should also discuss sharing policies with AI developers, whether they are from the same organization or from external institutions (e.g., co-developers or independent research groups). Open-access publications under various Creative Commons licenses allow for the broadest distribution of datasets (see the section below). To enable others to use the dataset effectively, usage notes (e.g., recommended software, suggested data splitting strategies) should be prepared. Before sharing data with other institutions, pseudonymization (or anonymization) of images and the removal of private information about the patient or owner—information that is not necessary for DL model development—should be performed.

## Open data

Considering the dependency of DL model development on available datasets, publishing datasets and making them accessible for research purposes is an extremely important contribution to science. The opportunities enabled by open data are diverse; some of the most significant are the following:

- Acceleration of methodological innovation: Open data allows researchers to explore innovative DL methods for the patterns of interest, fostering the development of state-of-the-art techniques and thereby gradually enhancing algorithmic capabilities.
- Augmentation of training data: Combining multi-institutional open with in-house datasets leads to increased quantity and variability (e.g., domain sources) of images needed for the development of robust and widely generalizable models.[96] These multi-institutional applicable models may facilitate research collaboration. Exclusive use of public data to develop a model for in-house application may not be appropriate due to laboratory specific image features.[10,50,56] Also, possible differences in the ground truth need to be considered.
- Broadening of performance evaluation: Open datasets can serve as secondary test sets, enabling more extensive evaluations of algorithmic robustness across diverse domains [96] and facilitating comparisons with previous studies. However, care must be taken to account for differences in ground truth definitions, as these may impair consistent performance evaluation across datasets.[23]
- Reduction of redundant efforts: By making datasets publicly available, researchers can avoid duplicating data collection efforts, saving time and resources. This allows them to focus on developing new methods or addressing novel research questions.
- Promotion of reproducibility of biomedical research: When DL-based image analysis is used to derive new biomedical insights, open data enhances the reproducibility of the analytic tools. In fact, a recent guideline for AI in life sciences defines open data as one of the minimum criteria for reproducibility.[48]

As listed in Table 4, some veterinary researchers have made their datasets publicly available for research purposes.[8,12,16,19,20,25,46,47,70,103,110-112] Some of these datasets span various domains, including several WSI-scanner types,[12,110] multiple tumor types,[12,111] multiple species,[12,70] and multiple laboratories,[12,70] showcasing collaborative initiatives that will foster robust DL model development. However, despite the growing number of publications on AI models for various applications in veterinary pathology, this literature review reveals that the majority of the veterinary research community remains hesitant to publish their datasets. While many authors do not publish their datasets, some have indicated in their manuscripts (via data availability statements) that they are willing to share data upon reasonable request and through individual agreements [40]. This hesitancy may stem from a desire to maintain control over proprietary data or plans to publish the data in subsequent studies.[68,70,93] While we recognize these restrictions faced by some AI developers, we hope that the opportunities provided by open data, as listed above, will positively influence the veterinary research community's willingness to publish datasets following the FAIR principles (Findable, Accessible, Interoperable, Reusable).[108]

An additional finding from this literature review (summarized in Table 4) is that there is a particular lack of datasets related to laboratory animals and toxicologic pathology applications. This gap likely reflects the legal and data privacy constraints faced by toxicologic companies. Supporting this observation, a survey of toxicologic pathologists revealed that most respondents do not consider sharing training datasets with co-developers for commercial or public projects.[82] Likewise, we acknowledge that other diagnostic companies are often restrictive in data sharing due to concerns over future commercial and proprietary interests.[6] Initiatives of whole slide imaging platforms through multi-institutional and multidisciplinary consortia, such as Bigpicture, are promising to overcome these legal and proprietary challenges of image availability and sharing.[79]

A variety of human datasets are also available,[12,96] however, also the majority of DL articles in human medicine either use existing open datasets or private datasets, and a request for more publicly accessible datasets is made.[104] Human datasets may serve as valuable supplements to animal datasets when developing models for veterinary applications. Previous research has demonstrated that models developed using human datasets (e.g., algorithms for mitotic figure detection in tumors) can be successfully transferred to animal samples, and vice-versa, while maintaining a high performance or archiving high performance after some transfer learning.[8,12,70,71] However, while cell and tissue morphology are often comparable between humans and animals, there may be significant differences, such as variations in the frequency of disease subtypes. These differences require the incorporation of species-specific data, particularly in test datasets.

When publishing datasets, it is essential to use data repositories that ensure long-term storage and access, such as Zenodo (https://zenodo.org/) or GitHub (https://github.com/). Clear usage notes and detailed dataset descriptions (e.g., included disease subtypes, digitization devices, etc.) should be provided to enable other researchers to utilize the data as intended. When working with public datasets, researchers should ideally adopt the original training, validation, and test partitions to facilitate comparison with published studies. The publication of images and their associated metadata requires a well-planned strategy to address ethical and data privacy considerations, even more so than for internal use of images.[40] For veterinary samples, ethical approval may be waived (depending on the regulatory framework of the individual country) when routine diagnostic samples are used. However, many journals require owner consent from animal owners, and diagnostic laboratories intending to use their caseload for future research should consider strategies for obtaining such consent (e.g., including a statement in the submission form).

**Table 4.** List of open-access datasets containing microscopic images of animal tissue samples and labels for patterns of interest. These datasets were developed for supervised deep learning (DL) model development. See Supplemental Table S2 for an expanded list. The datasets were identified through an extensive literature review of primary research articles; we acknowledge that some datasets may be available in open data repositories without a corresponding journal publication and therefore may have been missed in our search.

| Reference | Acronym | Tissue type | Species | Pattern of interest | Annotation type/shape |
|---|---|---|---|---|---|
| **Histology** | | | | | |
| Burrai et al. [25] | CMTD | Mammary tumors | Dog | Benign vs. malignant | Image classification |
| Bertram et al. [19] | MITOS_WSI_CCMCT | Cutaneous mast cell tumors | Dog | Mitotic figures | Spot annotations |
| Aubreville et al. [8] | MITOS_WSI_CMC | Mammary carcinoma | Dog | Mitotic figures | Spot annotations |
| Aubreville et al. [12] | MIDOG++ | Several tumor types | Dog, and human | Mitotic figures | Spot annotations |
| Banerjee et al. [16] | AtNorM-MD | Several tumor types | Dog and human | Normal vs. atypical mitotic figures | Patch classification |
| Bertram et al. [20] | N/A | Cutaneous mast cell tumors | Dog | Bi- and multinucleated tumor cells | Spot annotations |
| Haghofer et al. [47] | N/A | Cutaneous mast cell tumors | Dog | Neoplastic nuclei | Polygons |
| Haghofer et al. [46] | N/A | Lymphoma | Dog and cat | Neoplastic nuclei | Polygons |
| Wulcan et al. [112] | N/A | Lymphoma | Cat | Intestinal issue compartments and lymphocytes | Polygons |
| Wilm et al. [111] | CATCH | Several cutaneous tumor types | Dog | Tissue types | Polygons |
| Wilm et al. [110] | MC-SCC | Skin with squamous cell carcinoma | Dog | Tissue types | Polygons |
| **Cytology** | | | | | |
| Marzahl et al. [70] | N/A | bronchioalveolar lavage fluid | Horse, cat, human  Bounding box | Hemo-siderophages | Multi-class bounding boxed |
| Vogelbacher al. [103] | N/A | Blood smear | Avian | Cell types | Polygons |

N/A, not available

# Conclusion

Creation of large-scale and high-quality datasets for supervised DL in microscopic image analysis is a complex and resource-intensive process. Bias in selected imaged and created labels will be reflected in the model's generalization performance and one's ability to interpret performance metrics. This article has therefore highlighted critical considerations for dataset creation regarding 1) image selection, 2) annotation software, 3) annotation methods, and 4) further considerations. For image selection, not only the sheer number matters; it is also essential to capture the image variability expected in the application cases. For example, if the development dataset was digitized with one WSI scanner and the model is later applied to images from another scanner, performance may drop considerably. Active learning and synthetic data can help to increase the quantity or informativeness of training datasets and reduce the overall time investment; however, these methods should not be used to construct the test subset. The selected annotation software must provide all project-required features, including viewing support for specific file formats, online or offline application, and - if applicable - specific annotation tools such as guided screening and plug-ins for computer assistance. While annotations are often created by a single annotator, for complex patterns of interest (such as mitotic figures) there may be relevant visual and cognitive bias of annotators. Key criteria for annotation quality are the three "C"s: correctness, completeness, and consistency. Label bias of a single annotator may be reduced through multi-expert majority votes or computer assistance (such as by screening for missed candidates). To reduce the annotation workload, computer-assisted methods (algorithmic inference on unlabeled images and real-time algorithm–annotator interactions) or fully computerized methods (such as stain-registered transfer of computerized annotations) can be applied, with particular care to ensure high label quality for the test dataset. Further considerations before concluding dataset creation include data cleaning, storage and backup, documentation of dataset characteristics, and deciding on data sharing policies.

# Acknowledgements, Conflict of interest, Funding

**Acknowledgement**

The authors acknowledge the use of ChatGPT (OpenAI, GPT-4) for assistance with proofreading and improving the clarity of the manuscript. The authors take full responsibility for the content, accuracy, and interpretation of the work presented in this publication.

**Declaration of Conflicting Interests**

The author(s) declared no potential conflicts of interest with respect to the research, authorship, and/or publication of this article.

**Funding**

The author(s) disclosed receipt of the following financial support for the research, authorship, and/or publication of this article: CAB and VW acknowledge support from the Austrian Research Fund (FWF, project number: I 6555). MA acknowledges support by the Deutsche Forschungsgemeinschaft (DFG, project number: 520330054). JA acknowledges support by the Bavarian State Ministry for Science and the Arts (project FOKUS-TML).

# Standard operating procedure (SOP) for dataset creation

This SOP template was designed for the creation of datasets that will be used for supervised deep learning (DL) model development and performance evaluation. It outlines various considerations for all steps of dataset creation, starting from selecting the pathology task to ensuring long-term storage. This SOP was specifically designed for whole-slide images (WSI) as the image type, and modifications to this SOP may be necessary if the dataset is created using a different image type. Dataset creators need to make several decisions that best align with their project requirements. If the development of the deep learning-based algorithm is an interdisciplinary endeavor (involving pathologists, non-pathologist annotators such as students, computer scientists, laboratory technicians/managers), adequate communication and agreements should be established throughout the various steps of dataset creation.

1. Select the pattern of interest (pathology task) and the pattern recognition task (such as image classification, object detection, semantic or instance segmentation) and decide on the future use case (real-world application images) of the DL model. Subsequent steps will depend on these decisions.
2. Define case selection criteria and retrieve samples:
    A. Define the target domain for the image set: animal species, organs/tissue (normal background), disease entity, disease subgroups (such as tumor grades, early and late disease stage, tumor subtypes)
    B. List possible sources of tissue variability that can be expected in the target domain of the application cases: biological variability of the tissue, technical variability of slide and image creation (staining composition, special methods such as inking of surgical margins or decalcification, WSI scanner types and settings, scan artifacts such as out-of-focus scans, etc.) and decide to include (at representative proportion or oversample) or exclude corresponding samples.
    C. Consider ethical and legal regulations for the use of the images and patient metadata related to data privacy, security and usage consent.
        i. Experimental samples: Ethical approval obtained?
        ii. Routine diagnostic samples: Informed consensus by the owner obtained?
    D. Define the intended case / image number.
    E. Select potential cases and retrieve samples:
        i. Consider stratified sampling strategies (in particular for small image sets) to ensure that all relevant subgroups are included: representative frequency or oversampling of rare subgroups
        ii. Consider optimizing label class balance.
            a. For image classification tasks: consider including roughly equal number of foreground (pattern of interest) and background (other pattern) images.
            b. For object detection and segmentation tasks: equal proportion of foreground and background image parts is typically not achievable (for example mitotic figures are

rare events in tumor sections), and data balance needs to be achieved during training by sampling strategies and augmentation methods. Oversampling of cases with a high density of the pattern of interest may nevertheless be of advantage.
   3. Create the image set:
      A. Create images, either by using existing slides or by producing new slides.
         i. Use of existing glass slides (physical archive) or WSI from digital archive:
            a. Consider using samples from a longer inclusion period to encompass natural variability of slide and image creation.
            b. Consider using glass slides and/or WSI from multiple laboratories and produced with different scanners to account for technical variabilities, unless the derived DL model should only be applied to a controlled laboratory environment.
            c. Retrieve image creation protocols of the inclusion period (staining protocol, digitization device and settings, etc.).
         ii. Creation of new glass slides from tissue blocks with subsequent digitization:
            a. Consider creating slides in multiple batches (batch effect of sectioning and staining).
            b. Document the tissue preparation steps.
            c. Select and document the digitization devices (WSI scanner company and type), digitization settings (such as scan resolution, for z-stacks number and distance of scan levels, density of focus points, image compression) and, if applicable, image processing steps (such as conversion of the file format, color normalization etc.).
               a. Verify that the scan settings are appropriate to detect the pattern of interest.
               b. Ensure that the WSI format that can be read by your annotation software and software for DL model development.
      B. Conduct quality control of the WSI (computerized focus score and/or visual inspection) and remove unsuited images from the image set.
         i. Define image exclusion criteria (such as images with too low focus score; visual quality check criteria based on proportion out-of-focus areas, poor tissue quality, artifacts etc.).
         ii. Exclude images with very poor quality (i.e., they do not allow accurate/reproducible annotations) only if this does not represent a realistic use case for the algorithm.
      C. For WSI, decide if annotations should be made on entire WSI (high time investment) or should be restricted to regions of interest of these WSI.
         i. For regions of interest:
            a. Define the region selection criteria: random (including difficult regions and background tissue), "perfect" region

(intentionally excluding difficult regions), hotspot regions (region with highest density of objects of interest), combination thereof, or using active learning methods.
        b. Define size and shape of the region.
        c. Create annotations of the selected regions (for traceability).
        d. The region of interest can be provided to the annotator in the WSI (object annotations are made withing the region of interest annotation) or as a cropped, separate image.
    D. Consider the use of special methods of image selection that decreases the need for manual annotations:
        i. Active learning: start with an initial, small dataset and add image regions that are expected the benefit the model training the most. This method should not be used for test datasets and a sufficiently large test dataset should be set aside from the start.
        ii. Multi-scanner registered images: Each glass slide is digitized with multiple WSI scanners and the WSIs are registered (the images are aligned in order to have the same objects in the same location). Annotations will be created on one reference scanner image and can be spatially transferred to the corresponding images from other scanners.
            a. Conduct control that the annotation transfer was accurate.
        iii. Synthetic data: Synthetic data can be used (as a supplement to real images) for moderately sized or imbalanced datasets to generate more examples of the pattern of interest or to transfer the style of other domains (such as other scanners) to the dataset. Synthetic data should never be used for the test dataset.
    E. Name image files in a systematic manner, possibly avoiding direct case identification (pseudonymization) to adhere to data protection regulations.
        i. For pseudonymized images keep a list for re-identification with the laboratory system to allow to update metadata, when needed at a later time point.
    F. Collect all relevant metadata for each image in the set. This is needed to describe the dataset distribution, to conduct balanced, patient-level dataset splitting, and to conduct sub-analysis / stratified analysis of algorithmic performance (test set).
        i. Metadata may include species, patient ID, disease entity, disease subgroup, laboratory of origin, staining method, scanner type and settings etc.
    G. Make a back-up copy of the image set (WSI, possibly cropped regions of interest, and metadata) at a secure storage.
4. Select an annotation software:
    A. Decide for an offline or online application depending on the level of collaboration.
    B. The software needs to support the image file format, or the images needs to be converted to a different file format.

- C. The software needs to provide the needed annotation tools on the desired level (e.g., single click spot annotation for object detection or polygons for segmentation) and other relevant features (such as guides screening or plugins).
5. Plan the annotation process and create labels:
    - A. Define the annotations classes and select a reasonable annotation shape (such as spot annotation, bounding box or polygon) and the annotation granularity.
    - B. Identify the level of difficult for pathologists to annotate the pattern of interest (annotation bias) based on available literature or based on an own small pilot annotation study. This information is important to decide on advances annotation methods to ensure label quality.
    - C. Decide on the most appropriate annotation workflow balancing time investment with annotation bias / quality (correctness, consistency and completeness).
        - i. The minimal annotation method is a single pathologist per image
            - a. Ideally the same pathologists will annotate all images for highest consistency.
            - b. If the burden per pathologist must be reduced, multiple pathologists may be used that each annotate a different set of images. Ideally a small subset of images should be annotated by all pathologist to evaluate the inter-annotator consistency.
        - ii. Annotator recruitment method: decide on the required annotator qualification and degree of experience, based on the difficulty of the pathology task.
        - iii. Consider using advances annotation methods (beyond a single annotator).
            - a. Manual methods: multi-annotator majority vote, reevaluation of the annotations, co-registration with other staining / IHC to provide decision support.
            - b. Computer-assisted annotations:
                - a. Algorithmic inference on unlabeled images and subsequent annotator review. These tools aim at increasing label speed.
                - b. Real-time algorithmic modification during manual creation of annotations. These tools aim at increasing the granularity of annotations at high label efficiency.
                - c. Algorithmic inference on previously labeled images and subsequent annotator review (such as algorithmic missed candidate screening or annotation maps). These tools aim at increasing label quality.
            - c. Computerized annotations:
                - a. Special stain-registered transfer of computerized annotations: The special stain (such as immunohistochemistry or histochemical stain) is

used to generate the ground truth, which is subsequently transferred to the HE image. The HE image is used as input for DL model development.
           iv. Provide the annotators with appropriate hardware (e.g., computer mouse or stylus, large monitor etc.)
        D. Create clear annotations instructions:
            i. Describe the annotation class(es) and their morphological features. Explain how to distinguish the pattern of interest from other background patterns; ideally provide examples images of clear and borderline structures.
        E. Conduct annotator training and familiarization with annotation instructions and software.
        F. Conduct a pilot phase of the annotation workflow evaluating the technical feasibility and label quality (such as by cross-check by experienced pathologists or intra- and/or inter-annotator consistency). Provide the annotators with feedback regarding their label quality and evaluate if modifications in the annotation workflow and/or labeling instructions are needed. This will help to optimize the annotation workflow before fully committing to it and will save time in the long run needed for re-annotating the images.
            i. Are there issues with usability of the annotation software?
           ii. Are the annotations instructions clear? Do the annotation methods need to be optimized?
          iii. Does the annotator require further training?
        G. Create annotations.
        H. At regular intervals during the annotation process, make a back-up of the intermediate database at a save, short-term storage, possibly using version control.
        I. Conduct data cleaning of the dataset (i.e. detect and correct/remove errors) and final quality control of annotations.
            i. Ensure that annotations for all images are completed by all annotators.
           ii. Data cleaning: automated removal of obvious annotations errors (such as duplicate annotations of the same objects, false annotation shape used, too small annotations to realistically be an object of interest, etc.).
          iii. Consider developing a first algorithm to evaluate if the dataset allows development of a sufficiently good model (improves of the model can be done in later phases of development). This might give insight in particular challenging pattern, that are not sufficiently represented in the current dataset.
          iv. Discuss if additional annotation steps to improve annotation quantity and quality (such as computer-assisted detection of missed candidates or annotations maps to increase label consistency; see above) are necessary.
6. Prepare complete and accurate documentation:

A. Describe all steps taken to create the final dataset (see all bullet points above). Additionally, the reporting checklist by Elfer et al. [35] may be used to ensure complete reporting.
B. Conduct analysis of the final dataset.
   i. Case characteristics: number of included cases and annotations and their distribution across image subgroups.
      a. Possibly describe the patient demographics (breeds, age groups, etc.).
   ii. For datasets with multiple annotators, explore inter-annotator variability.
7. Prepare usage note for the dataset that allow future usage in the intended manner:
   A. Provide suggestions for dataset splitting into training, validation and test subsets, that can be consistently used across future studies (consistent partition is needed to allow comparison between studies).
   B. List recommended software for the use of the dataset.
8. Decide on data sharing policies:
   A. When dataset is shared with another institutions, apply pseudonymization (or anonymization) of images and removal of private metadata of the patient owners.
   B. Consider creating public accessibility to the dataset through data repositories and Creative Commons licensing.
9. Find a solution for secure long-term storage and back-up of the dataset.

# Supplemental Materials

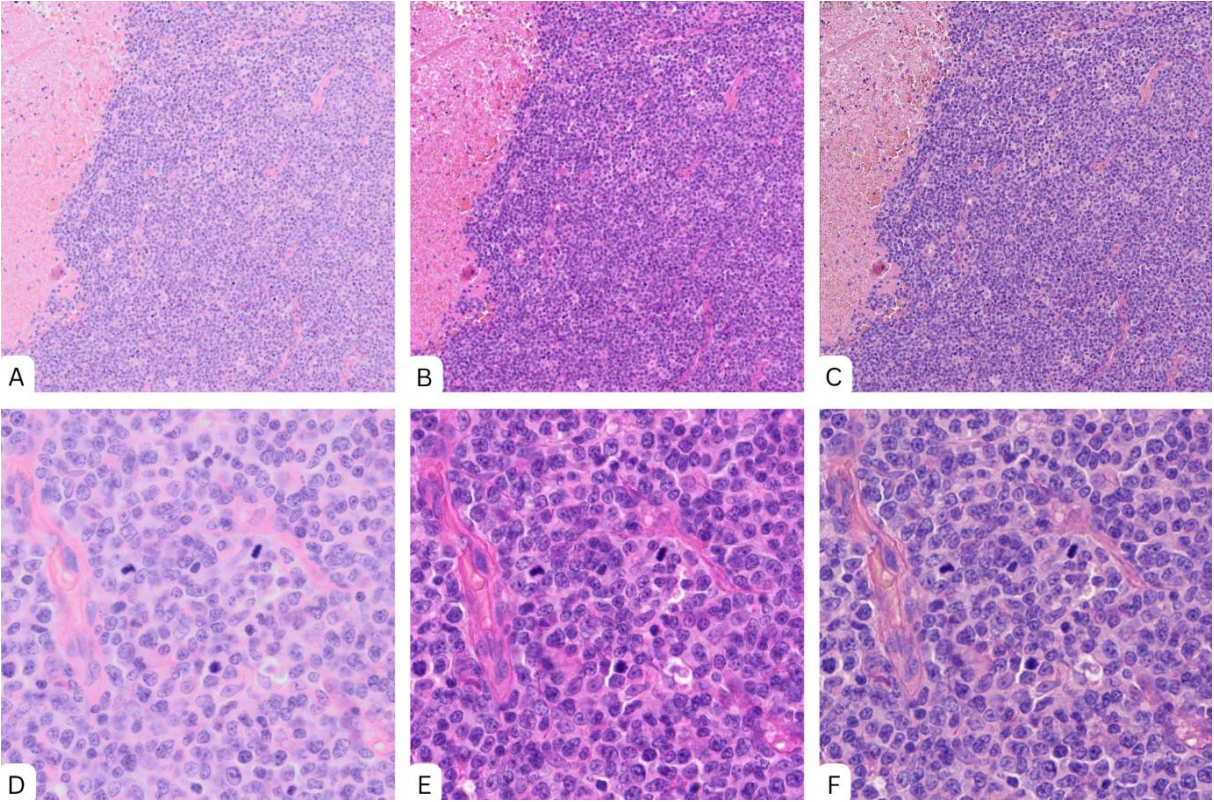

**Supplemental Figure S1.** Comparison of histological images from three different whole-slide image (WSI) scanners. All screenshots were taken on the same monitor using the same viewing software to avoid different color calibration. The upper row (**A-C**) shows images at low magnification and the lower row (**D-F**) shows images at higher magnification. **A, D:** Olympus VS 200, **B, E:** 3DHistech Pannoramic Scan I, **C, F:** Hamamatsu Nanozoomer S60

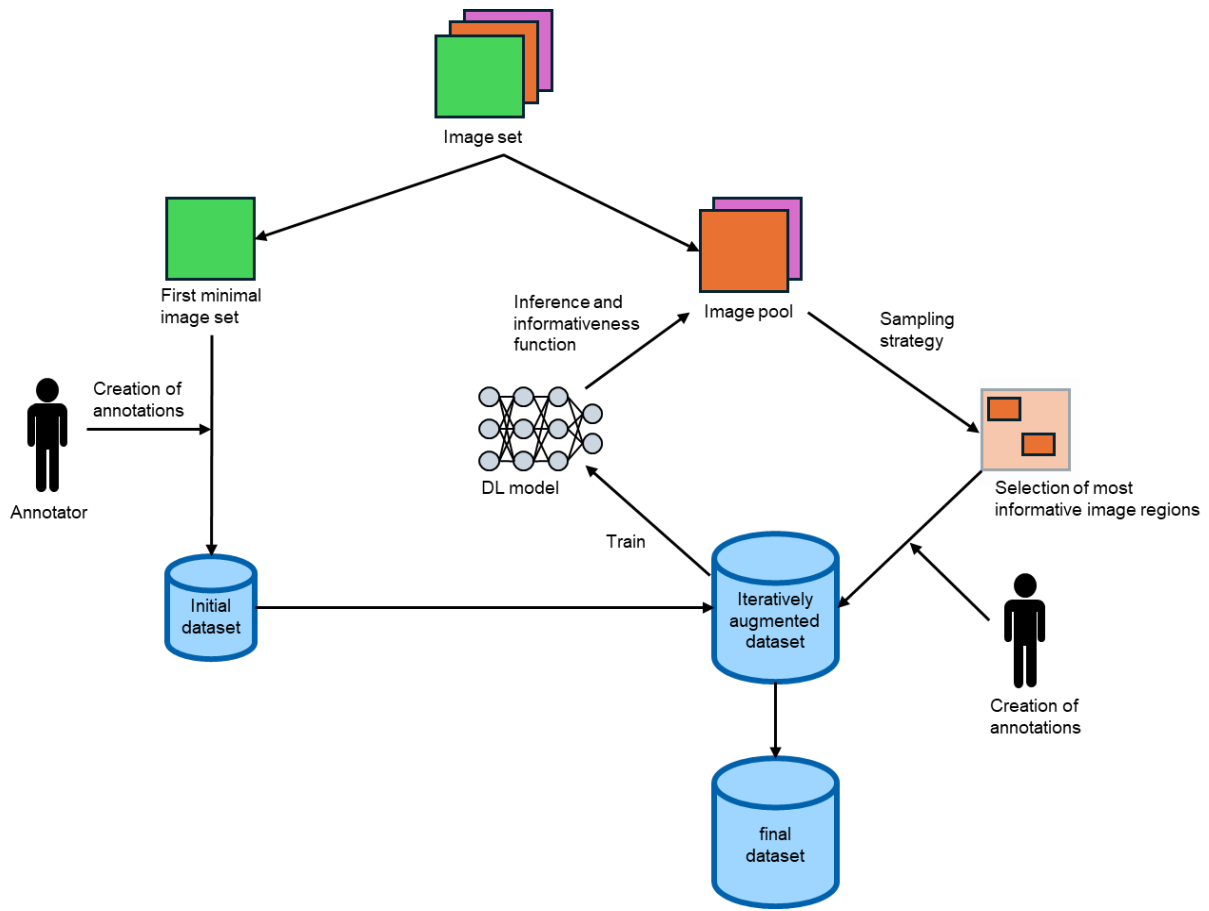

**Supplemental Figure S2.** Workflow of active learning that helps to reduce annotation effort for the training dataset by focusing on most meaningful image regions. An image set is created, from which a subset is used to create first annotations. Once an initial dataset has been created, a deep learning (DL) model can be trained to predict an informativeness function on new images from an image pool. Based on an informativeness metric and a dedicated sampling strategy, images can be selected that are assumed to be most informative for model training. For whole-slide images, typically smaller regions within the entire image are selected. Annotations are created for these selected images regions and added to the database. The iterative circle of adding images / image regions is terminated once the annotation budget is exhausted, or the intended model performance is achieved.

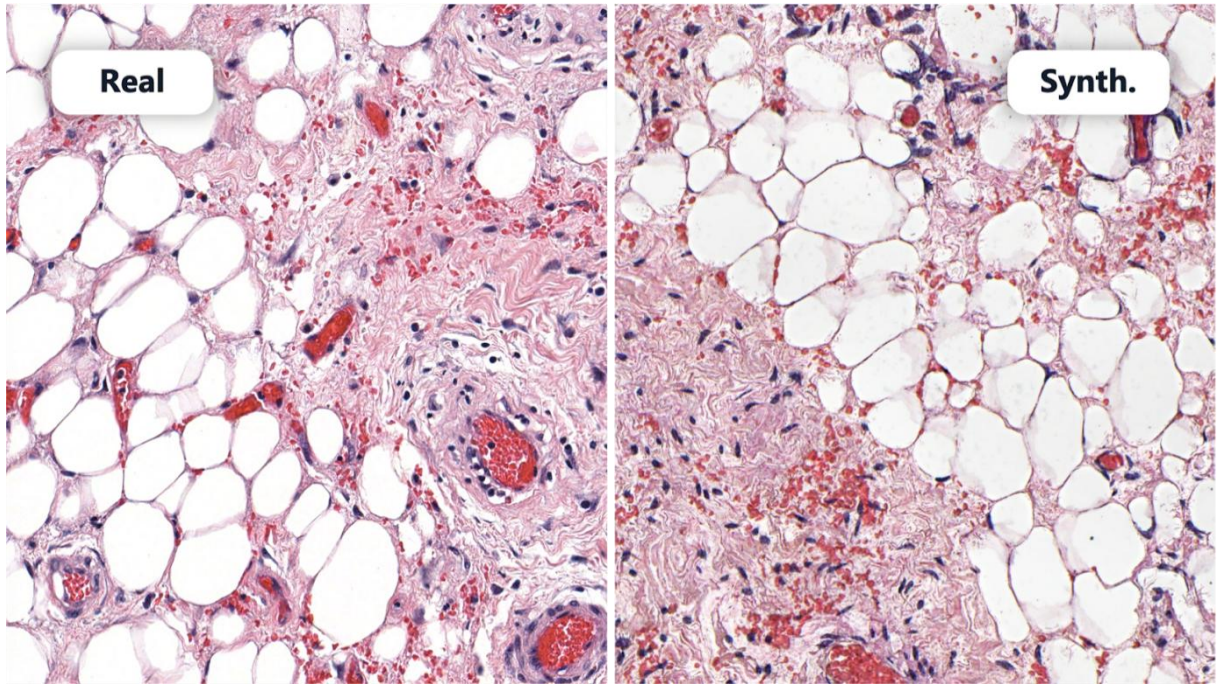

**Supplemental Figure S3.** Comparison between a real (left) and a synthetic (right) histopathology sample created with a controlled parameterized simulation technique as described by Mill et al.

Mill L, Aust O, Ackermann JA, et al. Deep learning-based image analysis in muscle histopathology using photo-realistic synthetic data. *Commun Med (Lond)*. 2025;5**:** 64. 10.1038/s43856-025-00777-y

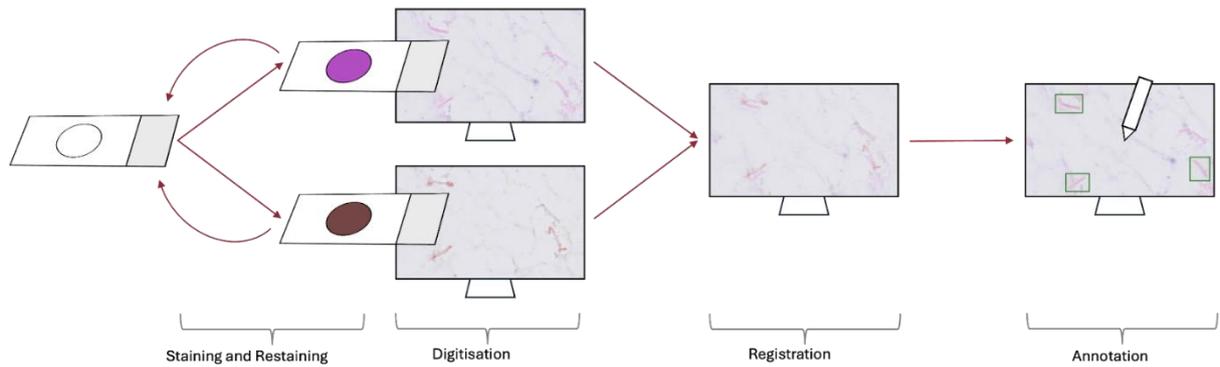

**Supplemental Figure S3.** Decision support for annotations in hematoxylin and eosin (HE) images by registration with specific staining method, using the example of helicobacter. The same histological section is stained with HE and immunohistochemistry (IHC), with scanning and destaining in between. Afterwards both digital images are registered (i.e., aligned on top of each other), the annotator can label objects in the H&E image based on a combined information provided in the HE and IHC images.

The order which staining method (HE or IHC) is done first can vary: 1) creating the HE slide first might impact the immunoreactivity of IHC and the HE image will be destroyed (thus is not available anymore, e.g. for scanning with further WSI scanners). 2) creating the IHC slide requires the use of a washable chromogen, may result in chromogen residues in the HE image, and may alter morphology of tissue and cells in the HE slide due to the more aggressive pretreatments (such as boiling) required for IHC.

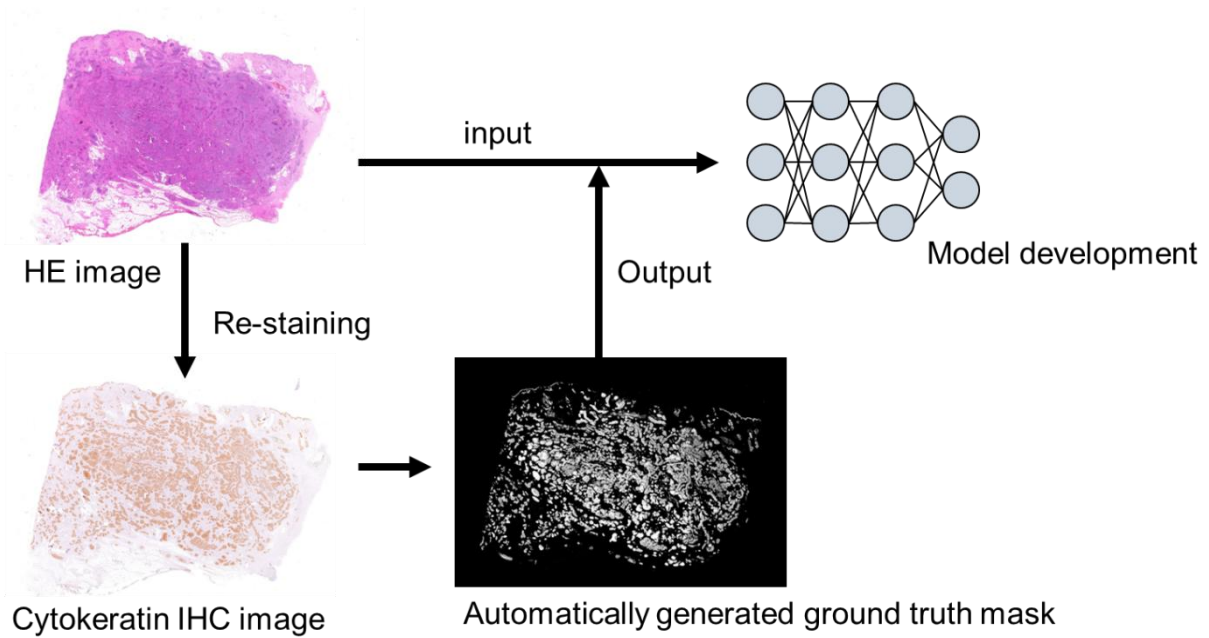

**Supplemental Figure S5.** Workflow for stain-registered transfer of computerized annotations using the example of neoplastic epithelium in a canine mammary carcinoma. The objective is to develop a deep learning model that can detect neoplastic epithelium in the HE image, while the ground truth is generated through immunohistochemistry (IHC). First, the HE slide is, after scanning, re-stained with IHC using an epithelial marker (in this case Cytokeratin AE1/3). From the IHC image a ground truth mask of the brown signal can be created using color deconvolution and filters. After registration of the HE and IHC images, the ground truth mask can be used as output for model training and testing, as the corresponding structures of both images are aligned to the same image locations.